\documentclass[10pt,twocolumn,letterpaper]{article}

\usepackage{iccv}
\usepackage{comment}
\usepackage{caption}
\usepackage{stfloats}
\usepackage{times}
\usepackage{epsfig}
\usepackage{graphicx}
\usepackage{amsmath}
\usepackage{amssymb}
\usepackage{ctable}
\usepackage{colortbl}
\usepackage{lipsum}
\usepackage{xcolor}
\usepackage{stfloats}

\usepackage{multirow}

\usepackage{subcaption}

\usepackage[accsupp]{axessibility}  

\usepackage[pagebackref=true,breaklinks=true,letterpaper=true,colorlinks,bookmarks=false]{hyperref}

\iccvfinalcopy 

\def\iccvPaperID{1398} 

\begin{document}

\newcommand{\OURS}{Pri3D}
\newcommand{\JI}[1]{\textbf{\textcolor{purple}{JI: #1}}}
\newcommand{\MATTHIAS}[1]{\textbf{\textcolor{red}{MATTHIAS: #1}}}
\newcommand{\ANGIE}[1]{\textbf{\textcolor{green}{Angie: #1}}}
\newcommand{\SAINING}[1]{\textbf{\textcolor{blue}{Saining: #1}}}
\newcommand{\BEN}[1]{\textbf{\textcolor{cyan}{Ben: #1}}}
\newcommand{\TODO}[1]{\textbf{\textcolor{red}{TODO: #1}}}

\definecolor{Gray}{gray}{0.92}
\definecolor{darkgreen}{rgb}{0.13, 0.55, 0.13}

\title{\OURS: Can 3D Priors Help 2D Representation Learning?}

\author{%
Ji Hou$^{1}$~~~~Saining Xie$^{2}$~~~~Benjamin Graham$^{2}$~~~~Angela Dai$^{1}$~~~~Matthias Nie{\ss}ner$^{1}$ \vspace{0.2cm}\\
$^{1}$Technical University of Munich~~~~$^{2}$Facebook AI Research
}

\twocolumn[{%
	\renewcommand\twocolumn[1][]{#1}%
	\maketitle
	\begin{center}
		\vspace{-0.35cm}
		\includegraphics[width=\linewidth]{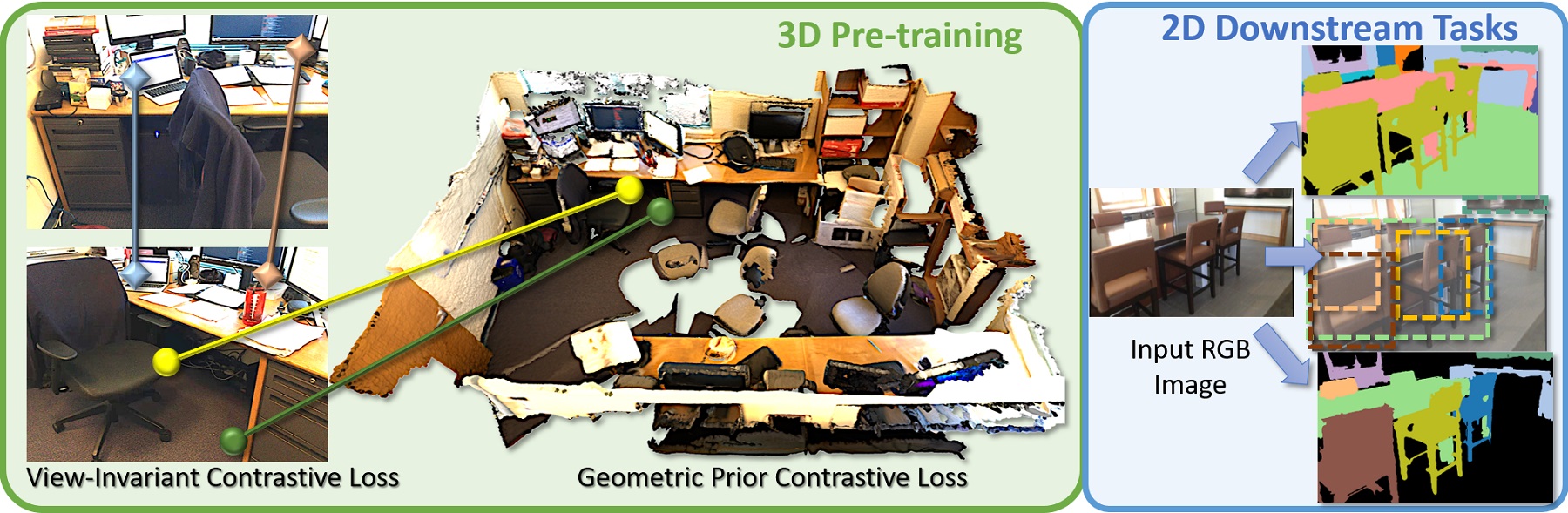}
		\captionof{figure}{
		\OURS{} leverages 3D priors for downstream 2D image understanding tasks: during pre-training, we incorporate view-invariant and geometric priors from color-geometry information given by RGB-D datasets, imbuing geometric priors into learned features. We show that these 3D-imbued learned features can effectively transfer to improved performance on 2D tasks such as semantic segmentation, object detection, and instance segmentation. 
		}
		\vspace{-0.05cm}
		\label{fig:teaser}
	\end{center}
}]

\maketitle

\begin{abstract}
Recent advances in 3D perception have shown impressive progress in understanding geometric structures of 3D shapes and even scenes.
Inspired by these advances in geometric understanding, we aim to imbue image-based perception with representations learned under geometric constraints. We introduce an approach to learn view-invariant, geometry-aware representations for network pre-training, based on multi-view RGB-D data, that can then be effectively transferred to downstream 2D tasks.
We propose to employ contrastive learning under both multi-view image constraints and image-geometry constraints to encode 3D priors into learned 2D representations. This results not only in improvement over 2D-only representation learning on the image-based tasks of semantic segmentation, instance segmentation and object detection on real-world indoor datasets, but moreover, provides significant improvement in the low data regime. We show significant improvement of 6.0\% on semantic segmentation on full data as well as 11.9\% on 20\% data against baselines on ScanNet. Our code is open sourced at \url{https://github.com/Sekunde/Pri3D}.
\end{abstract}
\section{Introduction}

In recent years, we have seen rapid progress in learning-based approaches for semantic understanding of 3D scenes, particularly in the tasks of 3D semantic segmentation, 3D object detection, and 3D semantic instance segmentation \cite{qi2017pointnet,dai20183dmv,thomas2019kpconv,huang2019texturenet,hou20193d,han2020occuseg,engelmann20203d,jiang2020pointgroup,voteNet}. 
Such approaches leverage geometric observations, exploiting the representation of points \cite{qi2017pointnet,qi2017pointnetplusplus}, voxels \cite{dai20183dmv,hou20193d}, or meshes \cite{huang2019texturenet} to obtain accurate 3D semantics.
These have shown significant promise towards realizing applications such as depth-based scene understanding for robotics, as well as augmented or virtual reality.
In parallel to the development of such methods, the availability of large-scale RGB-D datasets \cite{silberman2012indoor,hua2016scenenn,chang2017matterport3d,dai2017scannet}, has further accelerated the research in this area.

One advantage of learning directly in 3D in contrast to learning solely from 2D images is that methods operate in metric 3D space; hence, it is not necessary to learn view-dependent effects and/or projective mappings.
This allows training 3D neural networks from scratch in a relatively short time frame and typically requires a (relatively) small number of training samples; e.g., state-of-the-art 3D neural networks can be trained with around 1000 scenes from ScanNet.
Our main idea is to leverage these advantages in the form of 3D priors for image-based scene understanding.

Simultaneously, we have seen tremendous progress on representation learning in the image domain, mostly powered by the success of recent contrastive learning based methods~\cite{wu2018unsupervised,he2020momentum,chen2020simple,grill2020bootstrap,caron2020unsupervised}. The exploration in 2D representation learning heavily relies on the paradigm of instance discrimination, where different augmented copies of the same instance are drawn closer. Different invariances can be encoded from those low-level augmentations such as random cropping, flipping and scaling, as well as color jittering. However, despite the common belief that 3D view-invariance is an essential property for a capable visual system~\cite{marr1979computational}, there remains little study linking the 3D priors and 2D representation learning.
The goal of our work is to explore the combination of contrastive representation learning with 3D priors, and offer some preliminary evidence towards answering an important question: can 3D priors help 2D representation learning? 

To this end, we introduce \OURS, which aims to learn with 3D priors in a pre-training stage and subsequently use them as initialization for fine-tuning on image-based downstream tasks such as semantic segmentation, detection, and instance segmentation.
More specifically, we introduce geometric constraints to a contrastive learning scheme, which are enabled by multi-view RGB-D data that is readily available.
We propose to exploit geometric correlations through implicit multi-view constraints between different images through the correspondence of pixels which correspond to the same geometry, as well as explicit correspondence of geometric patches which correspond to image regions.
This imbues geometric knowledge into the learned representations of the image inputs which can then be leveraged as pre-trained features for various image-based vision tasks, particularly in the low training data regime.

We demonstrate our approach by pre-training on ScanNet~\cite{dai2017scannet} under these geometric constraints for representation learning, and show that such self-supervised pre-training (i.e., no semantic labels are used) results in improved performance on 2D semantic segmentation, instance segmentation and detection tasks.
We demonstrate this not only on ScanNet data, but also generalizing to improved performance on NYUv2~\cite{silberman2012indoor} semantic segmentation, instance segmentation and detection tasks. Moreover, leveraging such geometric priors for pre-training provides robust features which can consistently improve performance under a wide range of amount of training data available. While we focus on indoor scene understanding in this paper, we believe our results can shed light on the the paradigm of representation learning with 3D priors and open new opportunities towards more general 3D-aware image understanding.

\vspace{0.2cm}
In summary, our contributions are:
\vspace{-0.2cm}
\begin{itemize}
    \item A first exploration of the effect of 3D priors for 2D image understanding tasks, where we demonstrate the benefit of 3D geometric pre-training towards complex 2D perception such as semantic segmentation, object detection, and instance segmentation.
    ~\vspace{-0.2cm}
    \item A new pre-training approach based on 3D-guided view-invariant constraints and geometric priors from color-geometry correspondence, which learns features that can be transferred to 2D representations, complementing and improving image understanding across multiple datasets.
\end{itemize}

\section{Related Work}
\paragraph{3D Scene Understanding.}
Research in 3D scene understanding has recently been spurred forward with the introduction of larger-scale, real-world 3D scanned scene datasets \cite{armeni_cvpr16, dai2017scannet, chang2017matterport3d, geiger2013vision}.
We have seen notable progress in development of methods for semantic segmentation~\cite{qi2017pointnet, qi2017pointnetplusplus, thomas2019kpconv, wu2019pointconv,dai20183dmv, hu2020jsenet, lin2020fpconv, yan2020pointasnl, huang2019texturenet, zhang2020fusion}, object detection~\cite{song2014sliding,song2016deep,voteNet,qi2018frustum,imvotenet,zhang2020h3dnet,nie2020rfd}, and instance segmentation~\cite{hou20193d, yi2018gspn, yang2019learning,lahoud20193d,hou2020revealnet,engelmann20203d,han2020occuseg,jiang2020pointgroup} in 3D.
In particular, the introduction of sparse convolutional neural networks~\cite{graham20183d,choy20194d} have presented a computationally-efficient paradigm producing state-of-the-art results in such tasks.
Inspired by the developments in 3D scene understanding, we introduce learned geometric priors to representation learning for image-based vision tasks, leveraging a sparse convolutional backbone for 3D features used during pre-training. 

In the past year, we have also seen new developments in 3D representation learning.
PointContrast~\cite{xie2020pointcontrast} first showed that unsupervised, contrastive-based pre-training improves performance across various 3D semantic understanding tasks.
Hou et al.~\cite{hou2020exploring} introduces spatial context into 3D contrastive pre-training, resulting in improved performance in 3D limited annotation and data scenarios. Zhang et al.~\cite{zhang2021self} introduces a instance-discrimination-style pre-training approach that directly operates on depth frames. Our approach bridges these concepts into feature learning that can be transferred to 2D image understanding tasks.

~\vspace{-0.75cm}
\paragraph{2D Contrastive Representation Learning.}
Representation learning has driven significant efforts in deep learning; on the image domain, pre-training a network on a rich set of data has been shown to improve performance in fine-tuning for a smaller target dataset for various applications.
In particular, the contrastive learning framework~\cite{hadsell2006dimensionality} to learn representations from similar/dissimilar pairs of data has been demonstrated to show incredible promise \cite{oord2018representation,hjelm2018learning,wu2018unsupervised,he2020momentum,chen2020simple,chen2020improved,grill2020bootstrap,caron2020unsupervised}. Notably, using an instance discrimination task in which positive pairs are created with data augmentation, MoCo~\cite{he2020momentum} shows that unsupervised pre-training can surpass various supervised counterparts in detection and segmentation tasks, and SimCLR~\cite{chen2020simple} further reduces the gap to supervised pre-training in linear classifier performance.
Our approach leverages multi-view geometric information to augment contrastive learning and imbue robust geometric priors into learned feature representations.

~\vspace{-0.75cm}
\paragraph{Multi-Modality Learning} CLIP~\cite{radford2021learning} firstly proposes to train on images but with natural language supervision, and achieves significant results on zero-shot learning. BPNet~\cite{hu2021bidirectional} proposes a bidirectional projection module to mutually leverage 2D and 3D information for semantic segmentation task. 3D-to-2D Distillation~\cite{liu20213d} introduces additional 3D network in the training phase to embed 3D features for 2D semantic segmentation task. Existing works need to modify networks or add fusion modules in the training and/or inference phases. To this end, our method is more flexible as our pre-trained weights can be directly used like the ImageNet pre-trained model without any further modules or 3D/NLP data in the downstream tasks.

~\vspace{-0.75cm}
\paragraph{Correspondences Matching} Schmidt et al.~\cite{schmidt2016self} advocates a new approach to learning visual descriptors for dense correspondence estimation for the re-localization purpose, e.g., in the SLAM context. Schuster et al.~\cite{schuster2019sdc} presents a robust, unified descriptor network leveraging stacked dilated convolutions (SDC) for larger receptive field to better estimate dense pixel matching. HumanGPS~\cite{tan2021humangps} estimates dense correspondences between human images under arbitrary camera viewpoints and body poses. Existing works focus on 2D-2D correspondences matching problem itself. Our approach uses 2D-3D as well as 2D-2D view-invariant correspondences matching as pretext task to embed 3D priors for 2D downstream tasks.
\section{Learning Representations from 3D Priors}
\label{sec:method}

\begin{figure*}
	\centering
	\includegraphics[width=0.97\linewidth]{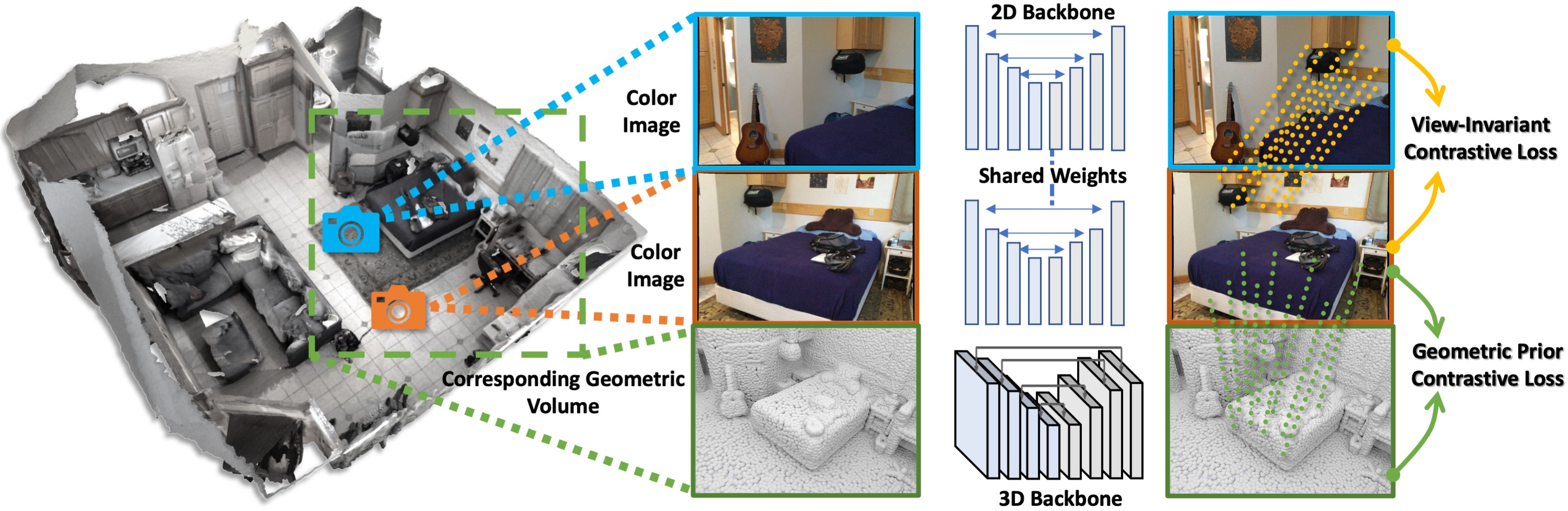}
	\vspace{-0.25cm}
	\caption{
	\textbf{Method Overview.} During pre-training, we use geometric constraints from RGB-D reconstructions to learn 3D priors for image-based representations. Specifically, we propose a contrastive learning formulation that models multi-view correspondences (View-Invariant Contrastive Loss) as well as geometry-to-image alignments (Geometric Prior Contrastive Loss). Our Pri3D pre-training strategy embeds geometric priors into the learned representations (in a form of pre-trained 2D convolutional network weights) that can be further leveraged for downstream 2D-only image understanding tasks.}
	\label{fig:method_overview}
	\vspace{-0.25cm}
\end{figure*}

In this section, we introduce \OURS; our key idea is to leverage constraints from RGB-D reconstructions, now readily available in various datasets \cite{geiger2013vision,song2015sun,dai2017scannet,chang2017matterport3d}, to embed 3D priors in image-based representations. 
From a dataset of RGB-D sequences, each sequence consists of depth and color frames, $\{D_i\}$ and $\{C_i\}$, respectively, as well as automatically-computed 6-DoF camera pose alignments $\{T_i\}$ (mapping from each camera space to world space) from state-of-the-art SLAM, all resulting in a reconstructed 3D surface geometry $S$.
We first revisit the simplest 3D priors, i.e. depth prediction~\cite{Hu2019RevisitingSI}. After seeing positive signals, we further explore more 3D priors.
Specifically, we observe that multi-view constraints can be exploited in order to learn view-invariance without the need of costly semantic labels.
In addition, we learn features through geometric representations given by the obtained geometry in RGB-D scans, again, without the need of human annotations.
For both, we use state-of-the-art contrastive learning in order to constrain the multi-modal input for training.
We show that these priors can be embedded in the image-based representations such that the learned features can be used as pre-trained features for purely image-based perception tasks; i.e., we can perform tasks such as image segmentation or instance segmentation on a single RGB image.
An overview of our approach is shown in Figure~\ref{fig:method_overview}.

\subsection{View-Invariant Learning}
\label{subsec:view_invariant}

In 2D constrative pre-training algorithms, a variety of data augmentations are used for finding positive matching pairs, such as MoCo~\cite{he2020momentum} and SimCLR~\cite{chen2020simple}. 
For instance, they use random crops as self-supervised constraints within the same image for positive pairs, and correspondences to crops from other images as negative pairs.
Our key idea is that with the availability of 3D data for training, we can leverage geometric knowledge to provide matching constraints between multiple images that see the same points.
To this end, we use the ScanNet RGB-D dataset~\cite{dai2017scannet} which provides a sequence of RGB-D images with camera poses computed by a state-of-the-art SLAM method~\cite{dai2017bundlefusion}, and reconstructed surface geometry $S$ \cite{niessner2013real}.
Note that both the pose alignments and the 3D reconstructions were obtained in a fully-automated fashion without any user input.

\begin{figure}[t]
\small
\centering
\includegraphics[width=1.0\linewidth]{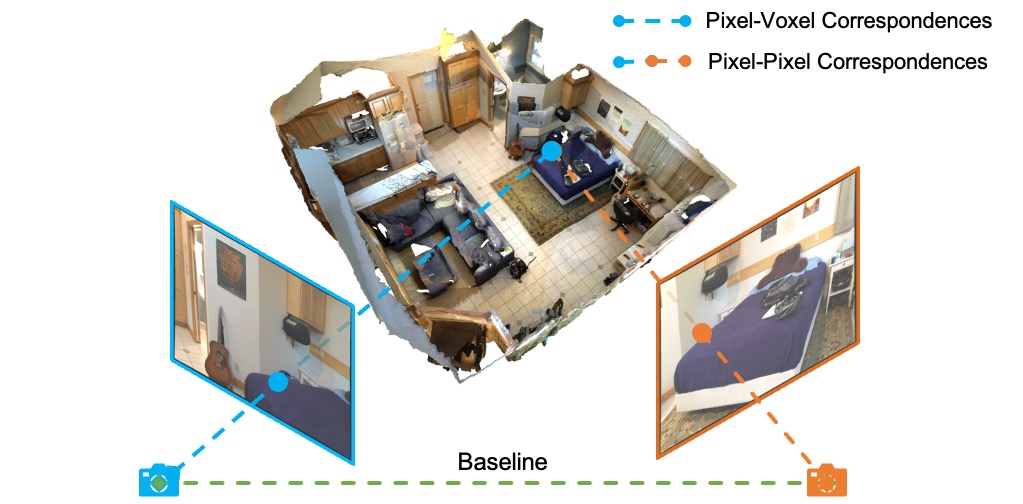}
\vspace{-0.5cm}
\caption{Illustration of finding correspondences between frames via epipolar geometry; world space as intermediary.}
\vspace{-0.5cm}
\label{fig:view_data}
\end{figure}

For a given RGB-D sequence in the train set, our method then leverages the 3D data to finding pixel-level correspondences between 2D frames.
We consider all pairs of frames $(i,j)$ from the RGB-D sequence.
We then back-project frame $i$'s depth map $D_i$ to camera space, and transform the points into world space by $T_i$.
The depth values of frame $j$ are similarly transformed into world space.
Pixel correspondences between the two frames are then determined as those whose 3D world locations lie within $2$cm of each other (see Figure~\ref{fig:view_data}).
We use the pairs of frames which have at least 30\% pixel overlap, with overlap computed as number of corresponding pixels in both frames divided by total number points in the two frames. 
In total, we sample around 840k pairs of images from the ScanNet training data. 

In the training phase, a pair of sampled images is input to a shared 2D network backbone.
In our experiments, we use a UNet-style~\cite{ronneberger2015u} backbone with ResNet~\cite{he2016deep} architecture as an encoder, but note that our method is agnostic to the underlying encoder backbone.
We then consider the feature map from decoder of the 2D backbone, where its size is half of the input resolution. 
For each image in the pair, we use the aforementioned pixel-to-pixel correspondences which refer to the same physical 3D point.
Note that these correspondences may have different color values due to view-dependent lighting effects but represent the same 3D world location; additionally, the regions surrounding the correspondences appear different due to different viewing angles.
In this fashion, we treat these pairs of correspondences as positive samples in contrastive learning; we use all non-matching pixels as negatives. Non-matching pixels are also defined within the set of correspondences. For a pair of frames with $n$ pairs of correspondences as positive samples, we use all $n(n-1)$ negative pairs (each of $n$ pixels from the first frame with each $n-1$ non-matching pixel from the second). Non-matching pixel-voxels are defined similarly but from a pair of frame and 3D chunk. 

Between the features of matching and non-matching pixel locations, we then compute a PointInfoNCE loss~\cite{xie2020pointcontrast}, which is defined as:
~\vspace{-0.2cm}
\begin{align}
\label{eq:pointinfonce}
    &\mathcal{L}_p = -\sum_{(a,b) \in M}\log\frac{\exp(\mathbf{f}_a\cdot\mathbf{f}_b/\tau)}{\sum_{(\cdot, k) \in M}\exp(\mathbf{f}_a\cdot\mathbf{f}_k/\tau)},
\end{align}
where $M$ is the set of pairs of pixel correspondences, and $f$ represents the associated feature vector of a pixel in the feature map.
By leveraging multi-view correspondences, we apply implicit 3D priors -- without any explicit 3D learning, we imbue view-invariance in the learned image-based features.

\subsection{Geometric Prior}
\label{subsec:geo_prior}

In addition to multi-view constraints,  we also leverage explicit geometry-color correspondences inherent to the RGB-D data during training.
For an RGB-D train sequence, the geometry-color correspondences are given by associating the surface reconstruction $S$ with the RGB frames of the sequence.
For each frame $i$, we compute its view frustum in the world space. 
A volumetric chunk $V_i$ of $S$ is then cropped from the axis-aligned bounding box of the view frustum.
We represent $V_i$ as a $2$cm resolution volumetric occupancy grid from the surface.
We thus consider pairs of color frames and geometric chunks $(C_i, V_i)$.

From the color-geometry pairs $(C_i, V_i)$, we compute pixel-voxel correspondences by projecting the depth values for each pixel in the corresponding frame $D_i$ into world space to find an associated occupied voxel in $V_i$ that lies within $2$cm of the 3D location of the pixel.

During training, we leverage the color-geometry correspondences with a 2D network backbone and a 3D network backbone. 
We use a UNet-style~\cite{ronneberger2015u} architecture with ResNet~\cite{he2016deep} encoder for the 2D network backbone, and a UNet-style sparse convolutional~\cite{graham20183d,choy20194d} 3D network backbone.
Similarly to view-invariant training, we also take the output from the decoder of 2D network backbone where its output size is half of the  input resolution. 
We then use the pixel-voxel correspondences in $(C_i, V_i)$ for contrastive learning, with positives as all matching pixel-voxel pairs and negatives as all non-matching pixel-voxel pairs.
We apply the PointInfoNCE loss (Equation~\ref{eq:pointinfonce}) with $f_i$ as the 2D features of a pixel, and $f_j$ is the feature vector from its 3D correspondence, and $M$  the set of 2D-3D pixel-voxel correspondence pairs.

\subsection{Joint Learning}
We can leverage not only the view-invariant constraints and geometric priors during training, but also learn jointly from the combination of both constraints.
We can thus employ a shared 2D network backbone and a 3D network backbone, with the 2D network backbone constrained by both view-invariant constraints and as the 2D part of the geometric prior constraint.

During training, we consider $(C_i, C_j, V_i, V_j)$ of overlapping color frames $C_i$ and $C_j$ as well as $V_i$ and $V_j$ which have geometric correspondence with $C_i$, $C_j$ respectively.
The shared 2D network backbone then processes $C_i, C_j$  and computes the view-invariant loss from Section~\ref{subsec:view_invariant}.
At the same time, $V_i$ and $V_j$ are processed by the 3D sparse convolutional backbone, with the loss (discussed in~\ref{subsec:geo_prior}) relative to the features of $C_i$ and $C_j$ respectively.
This embeds both constraints into the learned 2D representations.

\section{Experimental Setup}
\label{setup}
Our approach aims to embed 3D priors into the learned 2D representation by leveraging our view-invariant and geometric prior constraints. In this section, we introduce our detailed experimental setup for pre-training with an RGB-D dataset and fine-tuning on downstream 2D scene understanding tasks.

~\vspace{-0.7cm}
\paragraph{Architecture for Pre-training.}
As described in the previous section, our pre-training method leverages the pixel-to-pixel and geometry-to-color correspondences for view-invariant contrastive learning. The specific form of our pre-training objective requires a feature extractor capable of providing per-pixel or per-3D-point features for the backbone architecture, as the positive and negative matches are defined over 2D pixels or 3D locations. 

Our meta-architectures for both view-invariant constraints and geometric priors are U-Nets~\cite{ronneberger2015u} with residual connections. The encoder part of the U-Net is a standard ResNet. For view-invariant learning with 2D image inputs, we use ResNet18 or ResNet50 as encoders. The decoder part of the U-Net architecture consists of convolutional layers and bi-linear interpolation layers. For learning geometric priors from 3D volumetric occupancy input, we use sparse convolutions~\cite{graham20183d}, specifically a Residual U-Net-32 backbone implemented with MinkowskiEngine~\cite{choy20194d}, using a $2$cm voxel size. 

~\vspace{-0.7cm}
\paragraph{Stage I: Pri3D encoder initialization.}
We empirically found that for the pre-training phase, good initialization of the encoder network is critical to make learning robust. Instead of starting with random initialization, we initialize the encoder with network weights trained on ImageNet (\ie we pre-train the network for pre-training). The whole pipeline can be seen as a two-stage framework. We note that our method aims to improve the \emph{general} representation learning, thus is not tied to a specific learning paradigm (\eg supervised pre-training or self-supervised pre-training). From this perspective, we can leverage supervised pre-training of ResNet~\cite{he2016deep} encoders with ImageNet~\cite{deng2009imagenet} data for encoder initialization for pre-training. We name this model \textbf{Pri3D}. 

Although the use of a supervised ImageNet pre-trained initialization is a common practice, for completeness we also evaluate Pri3D in an unsupervised pipeline without using ImageNet labels. Results suggest that Pri3D does not rely on any semantic supervision (\eg ImageNet labels) to succeed, and still is able to achieve a substantial gain in this setup. We name this variant \textbf{Unsupervised Pri3D}. Further results of Unsupervised Pri3D are demonstrated in supplementary materials.

~\vspace{-0.7cm}
\paragraph{Stage II: Pri3D pre-training on ScanNet.}
Our pre-training method is enabled by the inherent geometry and color information present in the RGB-D data sequences. For pre-training, we leverage the color image and geometric reconstructions provided by the automatic reconstruction pipeline of ScanNet~\cite{dai2017scannet}; note that we do not use the semantic annotations during pre-training.
We also use a depth proxy loss by default in our experiments. Please refer to supplementary materials for details.
ScanNet contains 2.5M images from 1513 ScanNet train video sequences.
We regularly sample every $25^{\textrm{th}}$ frame without any other filtering (e.g., no control on viewpoint variation), and compute the set of overlapping pairs of frames that have $>30\%$ pixel overlap, resulting in $\approx 840$k frame pairs for which we compute their corresponding geometric chunks for each image, in order to apply both our view-invariant and geometric prior constraints.

~\vspace{-0.7cm}
\paragraph{Downstream Fine-tuning.} We evaluate our Pri3D models by fine-tuning them on a suite of downstream image-based scene understanding tasks. We use two datasets, ScanNet~\cite{dai2017scannet} and NYUv2~\cite{silberman2012indoor}, and the three tasks of semantic segmentation, object detection, and instance segmentation. As our pre-training dataset is ImageNet and ScanNet, fine-tuning on ScanNet represents a scenario of in-domain transfer---it would be interesting to know if the 3D priors can help with 2D representations for image-based tasks on \emph{the same dataset}. We further evaluate the performance of Pri3D on the NYUv2 dataset which maintains different statistics. This represents a out-of-domain transfer scenario. For semantic segmentation tasks, we directly use the U-Net architecture for dense prediction. The encoder and decoder networks are both pre-trained with Pri3D. For instance segmentation and detection tasks, we use Mask-RCNN~\cite{he2017mask} framework implemented in Detectron2~\cite{wu2019detectron2}. Only the backbone encoder part is pre-trained.

~\vspace{-0.7cm}
\paragraph{Implementation details.} 
For pre-training, we use an SGD optimizer with learning rate 0.1 and batch-size of 64. The learning rate is decreased by a factor of 0.99 every 1000 steps, and our method is trained for 60,000 iterations. For MoCoV2~\cite{chen2020improved}, we use the official PyTorch implementation. MoCoV2 is trained for 100 epochs with batch size 256. The fine-tuning experiments on semantic segmentation are trained with a batch size of 64 for 80 epochs. The initial learning rate is 0.01, with polynomial decay with power 0.9. All experiments are conducted on 8 NVIDIA V100 GPUs.

~\vspace{-0.7cm}
\paragraph{Baselines.} 
As we are using additional RGB-D data from ScanNet, it is important to benchmark our method against relevant baselines in order to answer the question: are 3D priors useful for 2D representation learning?
\begin{itemize}
\item \textbf{Supervised ImageNet Pre-training (IN)}. We use the ImageNet pre-trained weights provided in torchvision; this represents a widely adopted paradigm for image-based tasks. \emph{No ScanNet data is involved.}
\item \textbf{1-Stage MoCoV2 (MoCoV2-IN+SN)}. We train MoCoV2 on an expanded dataset that combines ImageNet with ScanNet. We explore two strategies: 1) Directly combining the two datasets with shuffled images and 2) mixing minibatches (sampling half images from ImageNet and the other half from ScanNet). In this case, \emph{we use ScanNet data but no 3D priors are considered.}
\item \textbf{2-Stage MoCoV2 (MoCoV2-supIN$\rightarrow$SN)}. As we use supervised pre-training (IN) in our method as encoder initialization, for fair comparison, we also try one version with (supervised) IN as the encoder initialization, then add another stage to fine-tune MoCoV2 with randomly shuffled ScanNet images. In this case, \emph{we use ScanNet data but no 3D priors are used.}
\item \textbf{Trivial Correspondences.} We use our framework but instead of learning from multi-view correspondences, we take one single-view image and create two copies by applying color space augmentations including: RGB jittering, random color dropping and Gaussian blur. Positive matches are defined on pixels at the same location. In this case, \emph{we use ScanNet data but no 3D priors are considered.}
\item \textbf{Depth Prediction.} We use the single frame depth prediction as a pre-training task. Note, our approach also leverages the depth prediction as a proxy loss by default. In this case, \emph{we use ScanNet data and a simple 3D prior is considered.} 
\end{itemize}

Through above baselines, we aim to justify that Pri3D learns to embed 3D priors in 2D representations that lead to an improved downstream performance; it is nontrivial to achieve the goal, given the auxiliary RGB-D dataset.

\section{Results}

\begin{figure*}
	\centering
	\includegraphics[width=0.97\linewidth]{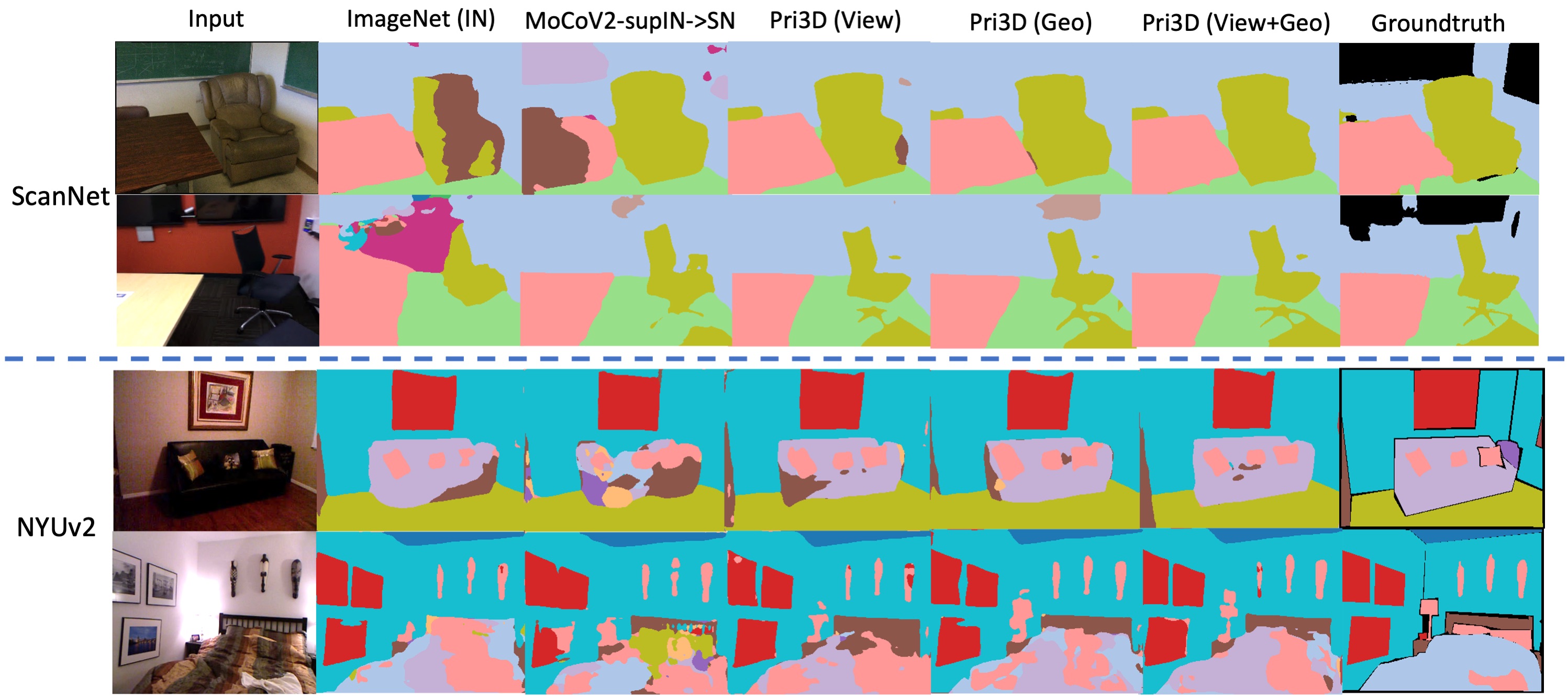}
	\caption{We show qualitative results on 2D semantic segmentation of ScanNet~\cite{dai2017scannet} and NYUv2~\cite{silberman2012indoor}. By encoding 3D priors, we obtain better segmentation results, in particular where when there are appearance variations over objects.}
	\label{fig:quality_result}
\end{figure*}

\label{results}
In this section, we present the results of our downstream fine-tuning results as well as relevant baselines mentioned in the previous section. 

\subsection{ScanNet}
We use our pre-trained network weights learned with Pri3D, and fine-tune for 2D semantic segmentation, object detection, and instance segmentation tasks on ScanNet~\cite{dai2017scannet} images, demonstrating the effectiveness of representation learning with 3D geometric priors. For fine-tuning, following the standard protocol in the ScanNet benchmark~\cite{dai2017scannet}: we sample every 100 2D frames, resulting in 20,000 train images and 5,000 validation images. 

\smallskip
\noindent \textbf{2D Semantic Segmentation.} 
We first show fine-tuning for semantic segmentation results in Table~\ref{tab:scannet_semseg_supervised}, in comparison with several baselines that also use ScanNet RGB-D data. We show the applicability of our approach with a standard ResNet50 backbone and a smaller ResNet18 backbone.

Comparing to just training the semantic segmentation model from scratch on downstream dataset ($39.1\%$ with ResNet50), all pre-training methods help significantly, even just using the ImageNet pre-training. This confirms the common belief in computer vision that a good 2D representation is essential for good performance on the target task. Several baselines, when adding the ScanNet RGB-D data, also works reasonably well, but \emph{not} much better than the naive ImageNet Pre-training baseline. This suggests that simply adding the ScanNet data into the representation learning pipeline does not necessarily lead to better results. Our Pri3D variants, including the view-invariant contrastive learning, geometry-color correspondence based contrastive learning and the combination of the two, provides substantially better representation quality that leads to improved semantic segmentation performance. We note that our method has a major performance boost ($+6.0\%$ absolute mIoU) even compared with the ImageNet Pre-training results. We believe this is an encouraging result and represents a practical use case as ImageNet pre-trained networks are often readily available.

\begin{figure}[h]
\small
\centering
\includegraphics[width=0.9\linewidth]{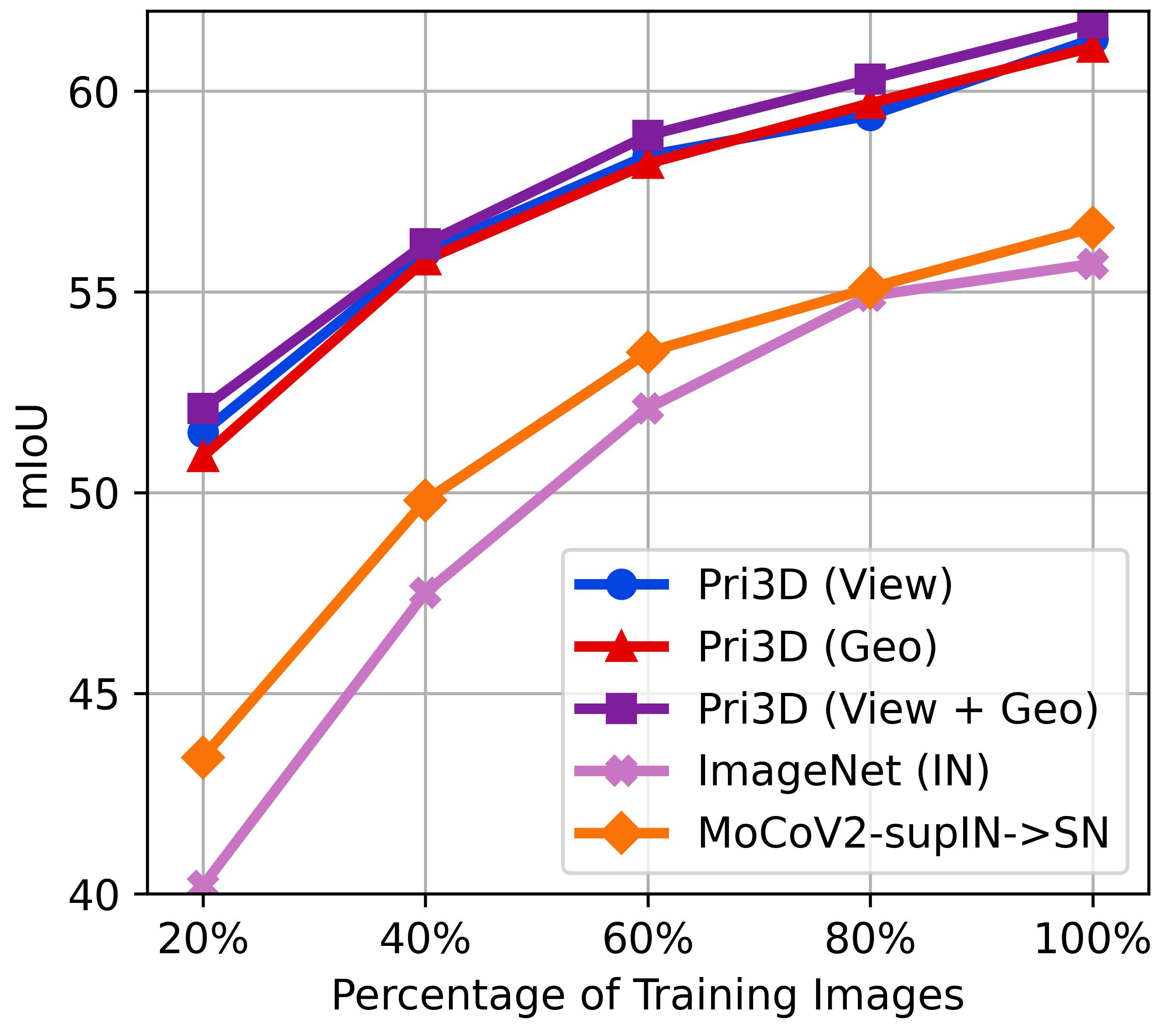}
~\vspace{-0.25cm}
\caption{\textbf{Data Efficient Learning on ScanNet (ResNet50 Backbone).} 
Using only 40\% of the training data, our pre-training can outperform supervised ImageNet pretraining when fine-tuned with 100\% data available for semantic segmentation. We see similar trends with a ResNet18 backbone, which is included in the appendix.
}
~\vspace{-0.5cm}
\label{fig:semseg_efficient50}
\end{figure}

Moreover, we evaluate our approach under limited data scenarios in Figure~\ref{fig:semseg_efficient50}. Our Pri3D pre-training shows an even larger gap when using a small subset of the training images, again compared to the strong ImageNet pre-training baseline. With only 20\% of the training data, we are able to recover $84\%$ and $80\%$ of the finetuning performance when using 100\% training data, with ResNet50 and ResNet18 backbone respectively.

\begin{table}[h]
  \centering
  \small
  \begin{tabular}{l|c|c}
  \specialrule{1.1pt}{0.1pt}{0pt}
Method & ResNet50  & ResNet18  \\
  \hline
   \textcolor{gray}{Scratch} & \textcolor{gray}{39.1} ~ & \textcolor{gray}{37.5} ~ \\
   ImageNet Pre-training (IN)                       &55.7 ~  &51.0 ~            \\ \hline
   MoCoV2-supIN→SN                                 &\quad 56.6~\tiny{\textcolor{gray}{(+0.9)}}              &\quad 52.9~\tiny{\textcolor{gray}{(+1.9)}}                 \\
   MoCoV2-IN+SN$_{\text{(combine)}}$        &\quad 54.9~\tiny{\textcolor{gray}{(-0.8)}}              &\quad -                                                    \\
   MoCoV2-IN+SN$_{\text{(mixing batch)}}$   &\quad 54.5~\tiny{\textcolor{gray}{(-1.2)}}              &\quad -                                                   \\
   Trivial Correspondences                      &\quad 56.4~\tiny{\textcolor{gray}{(+0.7)}}              &\quad 52.1~\tiny{\textcolor{gray}{(+1.1)}}                \\
   Depth Prediction       &\quad 58.4~\tiny{\textcolor{gray}{(+2.7)}} & - \\
  \hline
   Pri3D (View)                                 &\quad 61.3~\tiny{\textcolor{gray}{(+5.6)}}              &\quad 54.4~\tiny{\textcolor{gray}{(+3.4)}}                \\
   Pri3D (Geo)                                  &\quad 61.1~\tiny{\textcolor{gray}{(+5.4)}}              &\quad 55.3~\tiny{\textcolor{gray}{(+4.3)}}                \\
   Pri3D (View + Geo)                           &\quad \textbf{61.7~\tiny{\textcolor{darkgreen}{(+6.0)}}}   &\quad \textbf{55.7~\tiny{\textcolor{darkgreen}{(+4.7)}}}   \\
  \specialrule{1.1pt}{0.1pt}{0pt}
  \end{tabular}
  ~\vspace{-0.25cm}
  \caption{\textbf{2D Semantic Segmentation on ScanNet.} Fine-tuning with Pri3D pre-trained models leads to significantly improved results compared to ImageNet pre-training. Pri3D learns better representations with 3D priors and compares favorably with other baselines that also uses auxiliary RGB-D data. Please refer to Sec.~\ref{setup} for the detailed setup for those baselines. Metric is mean intersection-over-union (mIoU).}
  ~\vspace{-0.6cm}
\label{tab:scannet_semseg_supervised}
\end{table}

\smallskip
\noindent \textbf{2D Object Detection and Instance Segmentation.} 
To demonstrate that Pri3D is generalizable for different image-based tasks, we show results on fine-tuning for object detection in Table~\ref{tab:scannet_det} and instance segmentation in Table~\ref{tab:scannet_insseg}. For both tasks, we observe similar behavior to the semantic segmentation counterpart. All pre-training methods bring substantial improvement over training from scratch, but Pri3D models stand out and yield more gain compared to ImageNet Pre-training alone (+3.2\% and +2.8\% AP@0.5 for instance segmentation and detection, respectively). We note that for this set of experiments, we only transfer the encoder weights, discarding the decoder weights in the U-Net architecture for pre-training. This resembles similar practice in language domains (\eg BERT~\cite{devlin2018bert}) and shows that the main gain of Pri3D is better encoder representations.

\begin{table}[h]
  \centering
  \small
  \begin{tabular}{l|c|c|c}
  \specialrule{1.1pt}{0.1pt}{0pt}
   Method &  AP@0.5 & AP@0.75  & AP  \\
  \hline
   \textcolor{gray}{Scratch} &  \textcolor{gray}{32.7} ~ & \textcolor{gray}{17.7} ~ & \textcolor{gray}{16.9} ~ \\
   ImageNet (IN)         & 41.7 ~ & 25.9 ~ & 25.1 ~ \\
   MoCoV2-supIN→SN           & \quad 43.5~\tiny{\textcolor{gray}{(+1.8)}} &\quad 26.8~\tiny{\textcolor{gray}{(+0.9)}} &\quad 25.8~\tiny{\textcolor{gray}{(+0.7)}}\\
  \hline
   Pri3D (View)      & \quad 43.7~\tiny{\textcolor{gray}{(+2.0)}} &\quad 27.0~\tiny{\textcolor{gray}{(+1.1)}} &\quad 26.3~\tiny{\textcolor{gray}{(+1.2)}}\\
   Pri3D (Geo)      & \quad 44.2~\tiny{\textcolor{gray}{(+2.5)}} &\quad \textbf{27.6~\tiny{\textcolor{darkgreen}{(+1.7)}}} &\quad 26.6~\tiny{\textcolor{gray}{(+1.5)}}\\
   Pri3D (View+Geo) & \quad \textbf{44.5~\tiny{\textcolor{darkgreen}{(+2.8)}}} &\quad 27.4~\tiny{\textcolor{gray}{(+1.5)}} &\quad \textbf{26.6~\tiny{\textcolor{darkgreen}{(+1.5)}}}\\
  \specialrule{1.1pt}{0.1pt}{0pt}
  \end{tabular}
  ~\vspace{-0.25cm}
  \caption{\textbf{2D Detection on ScanNet.}  Fine-tuning with Pri3D pre-trained models leads to improved object detection results across different metrics compared to ImageNet pre-training and a strong MoCo-style pre-training method.}
  ~\vspace{-0.25cm}
\label{tab:scannet_det}
\end{table}

\begin{table}[h]
  \centering
  \small
  \begin{tabular}{l|c|c|c}
  \specialrule{1.1pt}{0.1pt}{0pt}
 Method &  AP@0.5 & AP@0.75  & AP  \\
  \hline
   \textcolor{gray}{Scratch} &  \textcolor{gray}{25.8} ~ & \textcolor{gray}{13.1} ~ &  \textcolor{gray}{12.2} ~ \\
   ImageNet (IN)          & 32.6 ~ & 17.8 ~ & 17.6 ~ \\
   MoCoV2-supIN→SN             &\quad 33.9~\tiny{\textcolor{gray}{(+1.3)}} &\quad 18.1~\tiny{\textcolor{gray}{(+0.3)}} & \quad 18.3~\tiny{\textcolor{gray}{(+0.7)}} \\
  \hline
   Pri3D (view)        &\quad 34.3~\tiny{\textcolor{gray}{(+1.7)}} &\quad 18.7~\tiny{\textcolor{gray}{(+0.9)}} & \quad 18.3~\tiny{\textcolor{gray}{(+0.7)}} \\
   Pri3D (geo).       &\quad 34.4~\tiny{\textcolor{gray}{(+1.8)}} &\quad 18.7~\tiny{\textcolor{gray}{(+0.9)}} & \quad 18.3~\tiny{\textcolor{gray}{(+0.7)}} \\
   Pri3D (view+geo) &\quad \textbf{35.8~\tiny{\textcolor{darkgreen}{(+3.2)}}} &\quad \textbf{19.3~\tiny{\textcolor{darkgreen}{(+1.5)}}} & \quad \textbf{18.7~\tiny{\textcolor{darkgreen}{(+1.1)}}} \\
  \specialrule{1.1pt}{0.1pt}{0pt}
  \end{tabular}
  ~\vspace{-0.25cm}
  \caption{\textbf{Instance Segmentation on ScanNet.} Fine-tuning with Pri3D pre-trained models leads to improved instance segmentation results compared to ImageNet pre-training and a strong MoCo-style pre-training method.}
  ~\vspace{-0.5cm}
\label{tab:scannet_insseg}
\end{table}

\vspace{-0.5cm}
\paragraph{SOTA Segmentation Network.} To demonstrate our method is agnostic to semantic segmentation backbones, we further show results with PSPNet and DeepLabV3/DeepLabV3+ in Table~\ref{tab:stoa_network}. Pri3D (Ours) consistently outperforms the baseline across different backbone choices.
\begin{table}[h]
\small
  \centering
  \begin{tabular}{l|c}
  \specialrule{1.1pt}{0.1pt}{0pt}
Method & ResNet50 \\
  \hline
   DeepLabV3 (ImageNet)  & 57.0 ~~ \\ 
   DeepLabV3 (Pri3D)     &\quad \textbf{61.3}~\tiny{\textcolor{darkgreen}{(+4.3)}}    \\ \hline
   DeepLabV3+ (ImageNet) & 57.8 ~~  \\ 
   DeepLabV3+ (Pri3D)   &\quad \textbf{61.6}~\tiny{\textcolor{darkgreen}{(+3.8)}}  \\ \hline
   PSPNet (ImageNet)     &59.7 ~~           \\ 
   PSPNet (Pri3D)        &\quad \textbf{62.8}~\tiny{\textcolor{darkgreen}{(+3.1)}}  \\ 
  \specialrule{1.1pt}{0.1pt}{0pt}
  \end{tabular}
  \vspace{-0.25cm}
  \caption{2D Semantic Segmentation on ScanNet (mIoU).}
  \vspace{-0.5cm}
\label{tab:stoa_network}
\end{table}

\subsection{NYUv2}
We show that our method learns transferable features across datasets. With Pri3D pre-trained on ScanNet RGB-D data, we explore fine-tuning on NYUv2~\cite{silberman2012indoor} for downstream 2D tasks.
The NYU-Depth V2 dataset is comprised of video sequences from a variety of indoor scenes, recorded by Microsoft Kinect RGB-D sensors. 
It contains 1449 densely labeled pairs of aligned RGB and depth images. We use the official split: 795 images for training, 654 images for test. Similar to ScanNet, we also evaluate on 3 popular downstream tasks, 2D semantic segmentation, object detection, and instance segmentation.
Table~\ref{tab:nyu_semseg} shows the semantic segmentation performance on NYUv2. 

We show the semantic segmentation fine-tuning performance on NYUv2 in Table~\ref{tab:nyu_semseg}; the object detection fine-tuning results in Table~\ref{tab:nyu_det}; and the instance segmentation fine-tuning results in Table~\ref{tab:nyu_insseg}. The experimental setup is similar to the ScanNet downstream fine-tuning counterpart, and we use supervised ImageNet pre-trained weights for encoder initialization of all methods. For all three tasks, we observe improved performance over different baselines such as training from scratch, training with ImageNet pre-trained weights, and MoCoV2-style pre-training on additional ScanNet data. Compared to the ImageNet pre-training baseline, we achieve a margin of $+4.4\%$ AP@0.5 for instance segmentation, $+4.8\%$ mIoU for semantic segmentation (ResNet50 backbone) and $+4.1\%$ AP@0.5 for object detection. 

\begin{table}[h]
  \centering
  \small
  \begin{tabular}{l|c|c}
  \specialrule{1.1pt}{0.1pt}{0pt}
Method & ResNet50  & ResNet18  \\
  \hline
  
   \textcolor{gray}{Scratch} & \textcolor{gray}{24.8} ~ & \textcolor{gray}{22.5} ~ \\
   ImageNet Pre-training (IN)       & 50.0 ~                                                 &  44.7 ~                                                      \\
   MoCoV2-supIN$\rightarrow$SN &\quad 47.6~\tiny{\textcolor{gray}{(-2.4)}}             &\quad 45.1~\tiny{\textcolor{gray}{(+0.4)}}                    \\
  \hline 
   Pri3D (View)                 &\quad 54.2~\tiny{\textcolor{gray}{(+4.2)}}              &\quad 48.2~\tiny{\textcolor{gray}{(+3.5)}}                    \\
   Pri3D (Geo)                  &\quad \textbf{54.8~\tiny{\textcolor{darkgreen}{(+4.8)}}}&\quad \textbf{48.6~\tiny{\textcolor{darkgreen}{(+3.9)}}}      \\
   Pri3D (View+Geo)                 &\quad 54.7~\tiny{\textcolor{gray}{(+4.7)}}              &\quad 48.1~\tiny{\textcolor{gray}{(+3.4)}}                   \\
  \specialrule{1.1pt}{0.1pt}{0pt}
  \end{tabular}
  \vspace{-0.25cm}
  \caption{\textbf{2D Semantic Segmentation on NYUv2}. Fine-tuning with Pri3D pre-trained models leads to improved semantic segmentation results compared to ImageNet pre-training and a strong MoCo-style pre-training method. Metric is Mean Intersection-Over-Union (mIoU).
  }
\label{tab:nyu_semseg}
\end{table}

\begin{table}[h]
  \centering
  \small
  \begin{tabular}{l|c|c|c}
  \specialrule{1.1pt}{0.1pt}{0pt}
Method & AP@0.5 & AP@0.75  & AP  \\
  \hline
   \textcolor{gray}{Scratch}                    & \textcolor{gray}{21.3} ~  & \textcolor{gray}{10.3} ~ & \textcolor{gray}{9.0} ~ \\
   ImageNet (IN)              & 29.9 ~ & 17.3 ~ & 16.8 ~ \\
  MoCoV2-supIN$\rightarrow$SN &\quad 30.1~\tiny{\textcolor{gray}{(+0.2)}} &\quad 18.1~\tiny{\textcolor{gray}{(+0.8)}} &\quad 17.3~\tiny{\textcolor{gray}{(+0.5)}} \\
  \hline
   Pri3D (View)               &\quad 33.0~\tiny{\textcolor{gray}{(+2.1)}} &\quad 19.8~\tiny{\textcolor{gray}{(+2.6)}} &\quad 18.9~\tiny{\textcolor{gray}{(+2.1)}}\\
   Pri3D (Geo)                &\quad 33.8~\tiny{\textcolor{gray}{(+2.9)}} &\quad 20.2~\tiny{\textcolor{gray}{(+2.9)}} &\quad 19.1~\tiny{\textcolor{gray}{(+2.3)}}\\
   Pri3D (View+Geo)           &\quad \textbf{34.0~\tiny{\textcolor{darkgreen}{(+4.1)}}} &\quad \textbf{20.4~\tiny{\textcolor{darkgreen}{(+3.1)}}} &\quad \textbf{19.4~\tiny{\textcolor{darkgreen}{(+2.6)}}} \\
  \specialrule{1.1pt}{0.1pt}{0pt}
  \end{tabular}
  \vspace{-0.25cm}
  \caption{\textbf{2D Object Detection on NYUv2.}  Better object detection AP can be obtained with Pri3D fine-tuning.}
  \vspace{-0.25cm}
\label{tab:nyu_det}
\end{table}

\begin{table}[h]
  \centering
  \small
  \begin{tabular}{l|c|c|c}
  \specialrule{1.1pt}{0.1pt}{0pt}
Method &  AP@0.5 & AP@0.75  & AP  \\
  \hline
  \textcolor{gray}{Scratch} & \textcolor{gray}{17.2} ~ & \textcolor{gray}{9.2} ~ & \textcolor{gray}{8.8} ~ \\
   ImageNet (IN)                    & 25.1 ~ & 13.9 ~ & 13.4 ~ \\
  MoCoV2-supIN$\rightarrow$SN       &\quad 27.2~\tiny{\textcolor{gray}{(+2.1)}}          &\quad 14.7~\tiny{\textcolor{gray}{(+0.2)}} &\quad 14.8~\tiny{\textcolor{gray}{(+1.4)}} \\
  \hline         
   Pri3D (View)                     &\quad 28.1~\tiny{\textcolor{gray}{(+3.0)}}          &\quad 15.7~\tiny{\textcolor{gray}{(+1.8)}} &\quad 15.7~\tiny{\textcolor{gray}{(+2.3)}} \\
   Pri3D (Geo)                      &\quad 29.0~\tiny{\textcolor{gray}{(+3.9)}}          &\quad 15.9~\tiny{\textcolor{gray}{(+2.0)}} &\quad 15.2~\tiny{\textcolor{gray}{(+1.8)}} \\
   Pri3D (View+Geo)                 &\quad \textbf{29.5~\tiny{\textcolor{darkgreen}{(+4.4)}}} &\quad \textbf{16.3~\tiny{\textcolor{darkgreen}{(+2.4)}}} &\quad \textbf{15.8~\tiny{\textcolor{darkgreen}{(+2.4)}}}  \\
  \specialrule{1.1pt}{0.1pt}{0pt}
  \end{tabular}
  \vspace{-0.25cm}
  \caption{\textbf{2D Instance Segmentation on NYUv2.} Better instance segmentation AP can be obtained with Pri3D.}
  \vspace{-0.25cm}
\label{tab:nyu_insseg}
\end{table}

\section{Conclusion}
We have introduced \OURS, a new method for representation learning for image-based scene understanding tasks. Our core idea is to incorporate 3D priors in a pre-training process whose constraints are applied under a contrastive loss formulation. We learn view-invariant and geometry-aware representations by leveraging multi-view and image-geometry correspondence from existing RGB-D dataset. We show that this results in significant improvement compared to 2D-only pre-training. With limited training data available, we outperform the semantic segmentation baselines by 11.9\% on ScanNet. We hope our results can shed light on the the general paradigm of representation learning with 3D priors and open up new opportunities towards 3D-aware image understanding.

{\small \noindent  \textbf{Acknowledgments} This work was supported by a TUM-IAS Rudolf Mo{\ss}bauer Fellowship, the ERC Starting Grant Scan2CAD (804724), the German Research Foundation (DFG) Grant Making Machine Learning on Static and Dynamic 3D Data Practical, a Google Research Grant, and the Bavarian State Ministry of Science and the Arts as coordinated by the Bavarian Research Institute for Digital Transformation (bidt).}

{\small
\bibliographystyle{ieee_fullname}
\bibliography{egbib}

\begin{thebibliography}{10}\itemsep=-1pt

\bibitem{armeni_cvpr16}
Iro Armeni, Ozan Sener, Amir~R. Zamir, Helen Jiang, Ioannis Brilakis, Martin
  Fischer, and Silvio Savarese.
\newblock {3D} semantic parsing of large-scale indoor spaces.
\newblock In {\em ICCV}, 2016.

\bibitem{caron2020unsupervised}
Mathilde Caron, Ishan Misra, Julien Mairal, Priya Goyal, Piotr Bojanowski, and
  Armand Joulin.
\newblock Unsupervised learning of visual features by contrasting cluster
  assignments.
\newblock In {\em NeurIPS}, 2020.

\bibitem{chang2017matterport3d}
Angel Chang, Angela Dai, Thomas Funkhouser, Maciej Halber, Matthias Niessner,
  Manolis Savva, Shuran Song, Andy Zeng, and Yinda Zhang.
\newblock Matterport3d: Learning from rgb-d data in indoor environments.
\newblock {\em arXiv preprint arXiv:1709.06158}, 2017.

\bibitem{chen2020simple}
Ting Chen, Simon Kornblith, Mohammad Norouzi, and Geoffrey Hinton.
\newblock A simple framework for contrastive learning of visual
  representations.
\newblock {\em ICML}, 2020.

\bibitem{chen2020improved}
Xinlei Chen, Haoqi Fan, Ross Girshick, and Kaiming He.
\newblock Improved baselines with momentum contrastive learning.
\newblock {\em arXiv preprint arXiv:2003.04297}, 2020.

\bibitem{choy20194d}
Christopher Choy, JunYoung Gwak, and Silvio Savarese.
\newblock {4D} spatio-temporal convnets: Minkowski convolutional neural
  networks.
\newblock In {\em CVPR}, 2019.

\bibitem{cordts2016cityscapes}
Marius Cordts, Mohamed Omran, Sebastian Ramos, Timo Rehfeld, Markus Enzweiler,
  Rodrigo Benenson, Uwe Franke, Stefan Roth, and Bernt Schiele.
\newblock The cityscapes dataset for semantic urban scene understanding.
\newblock In {\em CVPR}, 2016.

\bibitem{dai2017scannet}
Angela Dai, Angel~X Chang, Manolis Savva, Maciej Halber, Thomas Funkhouser, and
  Matthias Nie{\ss}ner.
\newblock Scannet: Richly-annotated {3D} reconstructions of indoor scenes.
\newblock In {\em CVPR}, 2017.

\bibitem{dai20183dmv}
Angela Dai and Matthias Nie{\ss}ner.
\newblock 3dmv: Joint 3d-multi-view prediction for 3d semantic scene
  segmentation.
\newblock In {\em Proceedings of the European Conference on Computer Vision
  (ECCV)}, pages 452--468, 2018.

\bibitem{dai2017bundlefusion}
Angela Dai, Matthias Nie{\ss}ner, Michael Zollh{\"o}fer, Shahram Izadi, and
  Christian Theobalt.
\newblock Bundlefusion: Real-time globally consistent 3d reconstruction using
  on-the-fly surface reintegration.
\newblock {\em ACM Transactions on Graphics (ToG)}, 36(4):1, 2017.

\bibitem{deng2009imagenet}
Jia Deng, Wei Dong, Richard Socher, Li-Jia Li, Kai Li, and Li Fei-Fei.
\newblock Imagenet: A large-scale hierarchical image database.
\newblock In {\em CVPR}, 2009.

\bibitem{devlin2018bert}
Jacob Devlin, Ming-Wei Chang, Kenton Lee, and Kristina Toutanova.
\newblock {BERT}: Pre-training of deep bidirectional transformers for language
  understanding.
\newblock In {\em NAACL}, 2019.

\bibitem{engelmann20203d}
Francis Engelmann, Martin Bokeloh, Alireza Fathi, Bastian Leibe, and Matthias
  Nie{\ss}ner.
\newblock {3D-MPA: Multi-Proposal Aggregation for 3D Semantic Instance
  Segmentation}.
\newblock In {\em CVPR}, 2020.

\bibitem{geiger2013vision}
Andreas Geiger, Philip Lenz, Christoph Stiller, and Raquel Urtasun.
\newblock Vision meets robotics: The kitti dataset.
\newblock {\em The International Journal of Robotics Research},
  32(11):1231--1237, 2013.

\bibitem{Geiger2012CVPR}
Andreas Geiger, Philip Lenz, and Raquel Urtasun.
\newblock Are we ready for autonomous driving? the kitti vision benchmark
  suite.
\newblock In {\em CVPR}, 2012.

\bibitem{graham20183d}
Benjamin Graham, Martin Engelcke, and Laurens van~der Maaten.
\newblock 3d semantic segmentation with submanifold sparse convolutional
  networks.
\newblock In {\em CVPR}, 2018.

\bibitem{grill2020bootstrap}
Jean-Bastien Grill, Florian Strub, Florent Altch{\'e}, Corentin Tallec, Pierre
  Richemond, Elena Buchatskaya, Carl Doersch, Bernardo Avila~Pires, Zhaohan
  Guo, Mohammad Gheshlaghi~Azar, et~al.
\newblock Bootstrap your own latent-a new approach to self-supervised learning.
\newblock {\em NeurIPS}, 2020.

\bibitem{hadsell2006dimensionality}
Raia Hadsell, Sumit Chopra, and Yann LeCun.
\newblock Dimensionality reduction by learning an invariant mapping.
\newblock In {\em 2006 IEEE Computer Society Conference on Computer Vision and
  Pattern Recognition (CVPR'06)}, volume~2, pages 1735--1742. IEEE, 2006.

\bibitem{han2020occuseg}
Lei Han, Tian Zheng, Lan Xu, and Lu Fang.
\newblock Occuseg: Occupancy-aware 3d instance segmentation.
\newblock In {\em CVPR}, 2020.

\bibitem{he2020momentum}
Kaiming He, Haoqi Fan, Yuxin Wu, Saining Xie, and Ross Girshick.
\newblock Momentum contrast for unsupervised visual representation learning.
\newblock In {\em CVPR}, 2020.

\bibitem{he2017mask}
Kaiming He, Georgia Gkioxari, Piotr Doll{\'a}r, and Ross Girshick.
\newblock Mask {R-CNN}.
\newblock In {\em ICCV}, 2017.

\bibitem{he2016deep}
Kaiming He, Xiangyu Zhang, Shaoqing Ren, and Jian Sun.
\newblock Deep residual learning for image recognition.
\newblock In {\em CVPR}, 2016.

\bibitem{hjelm2018learning}
R~Devon Hjelm, Alex Fedorov, Samuel Lavoie-Marchildon, Karan Grewal, Phil
  Bachman, Adam Trischler, and Yoshua Bengio.
\newblock Learning deep representations by mutual information estimation and
  maximization.
\newblock {\em ICLR}, 2019.

\bibitem{hou20193d}
Ji Hou, Angela Dai, and Matthias Nie{\ss}ner.
\newblock {3D-SIS: 3D Semantic Instance Segmentation of RGB-D Scans}.
\newblock In {\em CVPR}, 2019.

\bibitem{hou2020revealnet}
Ji Hou, Angela Dai, and Matthias Nie{\ss}ner.
\newblock {RevealNet: Seeing Behind Objects in RGB-D Scans}.
\newblock In {\em CVPR}, 2020.

\bibitem{hou2020exploring}
Ji Hou, Benjamin Graham, Matthias Nie{\ss}ner, and Saining Xie.
\newblock Exploring data-efficient 3d scene understanding with contrastive
  scene contexts.
\newblock In {\em CVPR}, 2021.

\bibitem{Hu2019RevisitingSI}
Junjie Hu, Mete Ozay, Yan Zhang, and Takayuki Okatani.
\newblock Revisiting single image depth estimation: Toward higher resolution
  maps with accurate object boundaries.
\newblock 2019.

\bibitem{hu2021bidirectional}
Wenbo Hu, Hengshuang Zhao, Li Jiang, Jiaya Jia, and Tien-Tsin Wong.
\newblock Bidirectional projection network for cross dimension scene
  understanding.
\newblock In {\em CVPR}, 2021.

\bibitem{hu2020jsenet}
Zeyu Hu, Mingmin Zhen, Xuyang Bai, Hongbo Fu, and Chiew-lan Tai.
\newblock Jsenet: Joint semantic segmentation and edge detection network for 3d
  point clouds.
\newblock {\em arXiv preprint arXiv:2007.06888}, 2020.

\bibitem{hua2016scenenn}
Binh-Son Hua, Quang-Hieu Pham, Duc~Thanh Nguyen, Minh-Khoi Tran, Lap-Fai Yu,
  and Sai-Kit Yeung.
\newblock Scenenn: A scene meshes dataset with annotations.
\newblock In {\em 2016 Fourth International Conference on 3D Vision (3DV)},
  pages 92--101. IEEE, 2016.

\bibitem{huang2019texturenet}
Jingwei Huang, Haotian Zhang, Li Yi, Thomas Funkhouser, Matthias Nie{\ss}ner,
  and Leonidas~J Guibas.
\newblock Texturenet: Consistent local parametrizations for learning from
  high-resolution signals on meshes.
\newblock In {\em Proceedings of the IEEE/CVF Conference on Computer Vision and
  Pattern Recognition}, pages 4440--4449, 2019.

\bibitem{jiang2020pointgroup}
Li Jiang, Hengshuang Zhao, Shaoshuai Shi, Shu Liu, Chi-Wing Fu, and Jiaya Jia.
\newblock {PointGroup: Dual-Set Point Grouping for 3D Instance Segmentation}.
\newblock In {\em CVPR}, 2020.

\bibitem{lahoud20193d}
Jean Lahoud, Bernard Ghanem, Marc Pollefeys, and Martin~R Oswald.
\newblock 3d instance segmentation via multi-task metric learning.
\newblock In {\em ICCV}, 2019.

\bibitem{MegaDepthLi18}
Zhengqi Li and Noah Snavely.
\newblock Megadepth: Learning single-view depth prediction from internet
  photos.
\newblock In {\em Computer Vision and Pattern Recognition (CVPR)}, 2018.

\bibitem{lin2014microsoft}
Tsung-Yi Lin, Michael Maire, Serge Belongie, James Hays, Pietro Perona, Deva
  Ramanan, Piotr Doll{\'a}r, and C~Lawrence Zitnick.
\newblock Microsoft coco: Common objects in context.
\newblock In {\em ECCV}, 2014.

\bibitem{lin2020fpconv}
Yiqun Lin, Zizheng Yan, Haibin Huang, Dong Du, Ligang Liu, Shuguang Cui, and
  Xiaoguang Han.
\newblock Fpconv: Learning local flattening for point convolution.
\newblock In {\em Proceedings of the IEEE/CVF Conference on Computer Vision and
  Pattern Recognition}, pages 4293--4302, 2020.

\bibitem{liu20213d}
Zhengzhe Liu, Xiaojuan Qi, and Chi-Wing Fu.
\newblock 3d-to-2d distillation for indoor scene parsing.
\newblock In {\em CVPR}, 2021.

\bibitem{marr1979computational}
David Marr and Tomaso Poggio.
\newblock A computational theory of human stereo vision.
\newblock {\em Proceedings of the Royal Society of London. Series B. Biological
  Sciences}, 204(1156):301--328, 1979.

\bibitem{nie2020rfd}
Yinyu Nie, Ji Hou, Xiaoguang Han, and Matthias Nie{\ss}ner.
\newblock Rfd-net: Point scene understanding by semantic instance
  reconstruction.
\newblock In {\em CVPR}, 2021.

\bibitem{niessner2013real}
Matthias Nie{\ss}ner, Michael Zollh{\"o}fer, Shahram Izadi, and Marc
  Stamminger.
\newblock Real-time 3d reconstruction at scale using voxel hashing.
\newblock {\em ACM TOG}, 32(6):169, 2013.

\bibitem{oord2018representation}
Aaron van~den Oord, Yazhe Li, and Oriol Vinyals.
\newblock Representation learning with contrastive predictive coding.
\newblock {\em arXiv preprint arXiv:1807.03748}, 2018.

\bibitem{imvotenet}
Charles~R Qi, Xinlei Chen, Or Litany, and Leonidas~J Guibas.
\newblock Imvotenet: Boosting {3D} object detection in point clouds with image
  votes.
\newblock In {\em CVPR}, 2020.

\bibitem{voteNet}
Charles~R. Qi, Or Litany, Kaiming He, and Leonidas~J. Guibas.
\newblock Deep hough voting for 3d object detection in point clouds.
\newblock {\em ICCV}, 2019.

\bibitem{qi2018frustum}
Charles~R Qi, Wei Liu, Chenxia Wu, Hao Su, and Leonidas~J Guibas.
\newblock Frustum pointnets for 3d object detection from rgb-d data.
\newblock In {\em CVPR}, 2018.

\bibitem{qi2017pointnet}
Charles~R Qi, Hao Su, Kaichun Mo, and Leonidas~J Guibas.
\newblock Pointnet: Deep learning on point sets for 3d classification and
  segmentation.
\newblock {\em CVPR}, 2017.

\bibitem{qi2017pointnetplusplus}
Charles~R Qi, Li Yi, Hao Su, and Leonidas~J Guibas.
\newblock Pointnet++: Deep hierarchical feature learning on point sets in a
  metric space.
\newblock {\em NeurIPS}, 2017.

\bibitem{radford2021learning}
Alec Radford, Jong~Wook Kim, Chris Hallacy, Aditya Ramesh, Gabriel Goh,
  Sandhini Agarwal, Girish Sastry, Amanda Askell, Pamela Mishkin, Jack Clark,
  et~al.
\newblock Learning transferable visual models from natural language
  supervision.
\newblock {\em arXiv preprint arXiv:2103.00020}, 2021.

\bibitem{ronneberger2015u}
Olaf Ronneberger, Philipp Fischer, and Thomas Brox.
\newblock U-net: Convolutional networks for biomedical image segmentation.
\newblock In {\em MICCAI}, 2015.

\bibitem{schmidt2016self}
Tanner Schmidt, Richard Newcombe, and Dieter Fox.
\newblock Self-supervised visual descriptor learning for dense correspondence.
\newblock In {\em ICRA}, 2017.

\bibitem{schuster2019sdc}
Ren{\'e} Schuster, Oliver Wasenmuller, Christian Unger, and Didier Stricker.
\newblock Sdc-stacked dilated convolution: A unified descriptor network for
  dense matching tasks.
\newblock In {\em CVPR}, 2019.

\bibitem{silberman2012indoor}
Nathan Silberman, Derek Hoiem, Pushmeet Kohli, and Rob Fergus.
\newblock Indoor segmentation and support inference from {RGB-D} images.
\newblock {\em ECCV}, 2012.

\bibitem{song2015sun}
Shuran Song, Samuel~P Lichtenberg, and Jianxiong Xiao.
\newblock {SUN RGB-D: A RGB-D Scene Understanding Benchmark Suite}.
\newblock In {\em CVPR}, 2015.

\bibitem{song2014sliding}
Shuran Song and Jianxiong Xiao.
\newblock Sliding shapes for 3d object detection in depth images.
\newblock In {\em ECCV}. 2014.

\bibitem{song2016deep}
Shuran Song and Jianxiong Xiao.
\newblock Deep sliding shapes for amodal 3d object detection in rgb-d images.
\newblock In {\em CVPR}, 2016.

\bibitem{tan2021humangps}
Feitong Tan, Danhang Tang, Mingsong Dou, Kaiwen Guo, Rohit Pandey, Cem Keskin,
  Ruofei Du, Deqing Sun, Sofien Bouaziz, Sean Fanello, et~al.
\newblock Humangps: Geodesic preserving feature for dense human
  correspondences.
\newblock {\em CVPR}, 2021.

\bibitem{thomas2019kpconv}
Hugues Thomas, Charles~R Qi, Jean-Emmanuel Deschaud, Beatriz Marcotegui,
  Fran{\c{c}}ois Goulette, and Leonidas~J Guibas.
\newblock Kpconv: Flexible and deformable convolution for point clouds.
\newblock In {\em CVPR}, 2019.

\bibitem{wu2019pointconv}
Wenxuan Wu, Zhongang Qi, and Li Fuxin.
\newblock Pointconv: Deep convolutional networks on 3d point clouds.
\newblock In {\em CVPR}, 2019.

\bibitem{wu2019detectron2}
Yuxin Wu, Alexander Kirillov, Francisco Massa, Wan-Yen Lo, and Ross Girshick.
\newblock Detectron2.
\newblock \url{https://github.com/facebookresearch/detectron2}, 2019.

\bibitem{wu2018unsupervised}
Zhirong Wu, Yuanjun Xiong, Stella~X Yu, and Dahua Lin.
\newblock Unsupervised feature learning via non-parametric instance
  discrimination.
\newblock In {\em CVPR}, 2018.

\bibitem{xie2020pointcontrast}
Saining Xie, Jiatao Gu, Demi Guo, Charles~R Qi, Leonidas~J Guibas, and Or
  Litany.
\newblock Pointcontrast: Unsupervised pre-training for 3d point cloud
  understanding.
\newblock {\em ECCV}, 2020.

\bibitem{yan2020pointasnl}
Xu Yan, Chaoda Zheng, Zhen Li, Sheng Wang, and Shuguang Cui.
\newblock Pointasnl: Robust point clouds processing using nonlocal neural
  networks with adaptive sampling.
\newblock In {\em Proceedings of the IEEE/CVF Conference on Computer Vision and
  Pattern Recognition}, pages 5589--5598, 2020.

\bibitem{yang2019learning}
Bo Yang, Jianan Wang, Ronald Clark, Qingyong Hu, Sen Wang, Andrew Markham, and
  Niki Trigoni.
\newblock Learning object bounding boxes for 3d instance segmentation on point
  clouds.
\newblock In {\em NeurIPS}, 2019.

\bibitem{yi2018gspn}
Li Yi, Wang Zhao, He Wang, Minhyuk Sung, and Leonidas Guibas.
\newblock {GSPN}: Generative shape proposal network for {3D} instance
  segmentation in point cloud.
\newblock In {\em CVPR}, 2019.

\bibitem{zhang2020fusion}
Jiazhao Zhang, Chenyang Zhu, Lintao Zheng, and Kai Xu.
\newblock Fusion-aware point convolution for online semantic 3d scene
  segmentation.
\newblock In {\em Proceedings of the IEEE/CVF Conference on Computer Vision and
  Pattern Recognition}, pages 4534--4543, 2020.

\bibitem{zhang2021self}
Zaiwei Zhang, Rohit Girdhar, Armand Joulin, and Ishan Misra.
\newblock Self-supervised pretraining of 3d features on any point-cloud.
\newblock {\em arXiv preprint arXiv:2101.02691}, 2021.

\bibitem{zhang2020h3dnet}
Zaiwei Zhang, Bo Sun, Haitao Yang, and Qixing Huang.
\newblock H3dnet: 3d object detection using hybrid geometric primitives.
\newblock In {\em European Conference on Computer Vision}, pages 311--329.
  Springer, 2020.

\end{thebibliography}
}

\clearpage
\noindent \textbf{\Large Appendix} \\ 
\begin{appendix}
In this appendix, we show data-efficient learning results on more downstream tasks spanning different percentages of used training images in Section~\ref{sec:data_efficient_learning}. We show more visualizations of semantic segmentation results in Section~\ref{sec:vis}. To provide more results across datasets, we further demonstrate a variety of experimental results of outdoor data in Section~\ref{sec:generalization}. As promised in main paper, we show the results of \textbf{Unsupervised Pri3D} in Section~\ref{sec:unpri3d}. In the end, we discuss the current limitations and potential directions of improvements in Section~\ref{sec:improvements}. 

\section{Data-Efficient Learning}\label{sec:data_efficient_learning}
\begin{figure}[h]
\small
\centering
\includegraphics[width=0.9\linewidth]{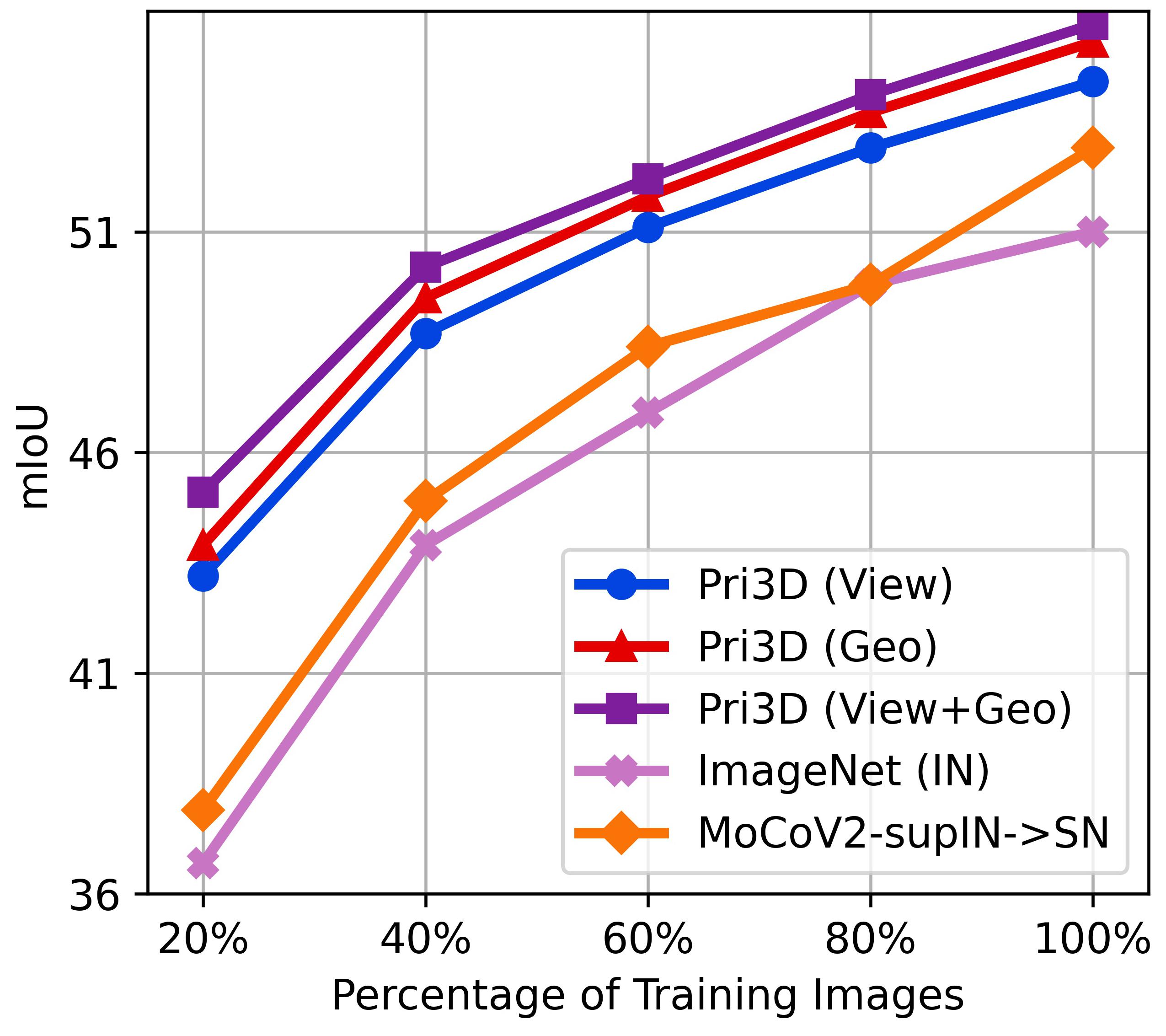}
\caption{\textbf{Data Efficient Learning on ScanNet 2D Semantic Segmentation Task.} Our method achieves consistently better results with a ResNet18 backbone.}
\label{fig:semseg_efficient18}
\end{figure}

We plot the curves across different percentages of used training data for data-efficient learning similarly to~\cite{hou2020exploring}. 
We demonstrate that our pre-training algorithm generalizes with different backbones. We show data-efficient learning curves on semantic segmentation task in ScanNet~\cite{dai2017scannet} with a ResNet18~\cite{he2016deep} backbone in Figure~\ref{fig:semseg_efficient18}. Please refer to main paper for results with a ResNet50 backbone.

\begin{figure}[h]
\small
\centering
\includegraphics[width=0.9\linewidth]{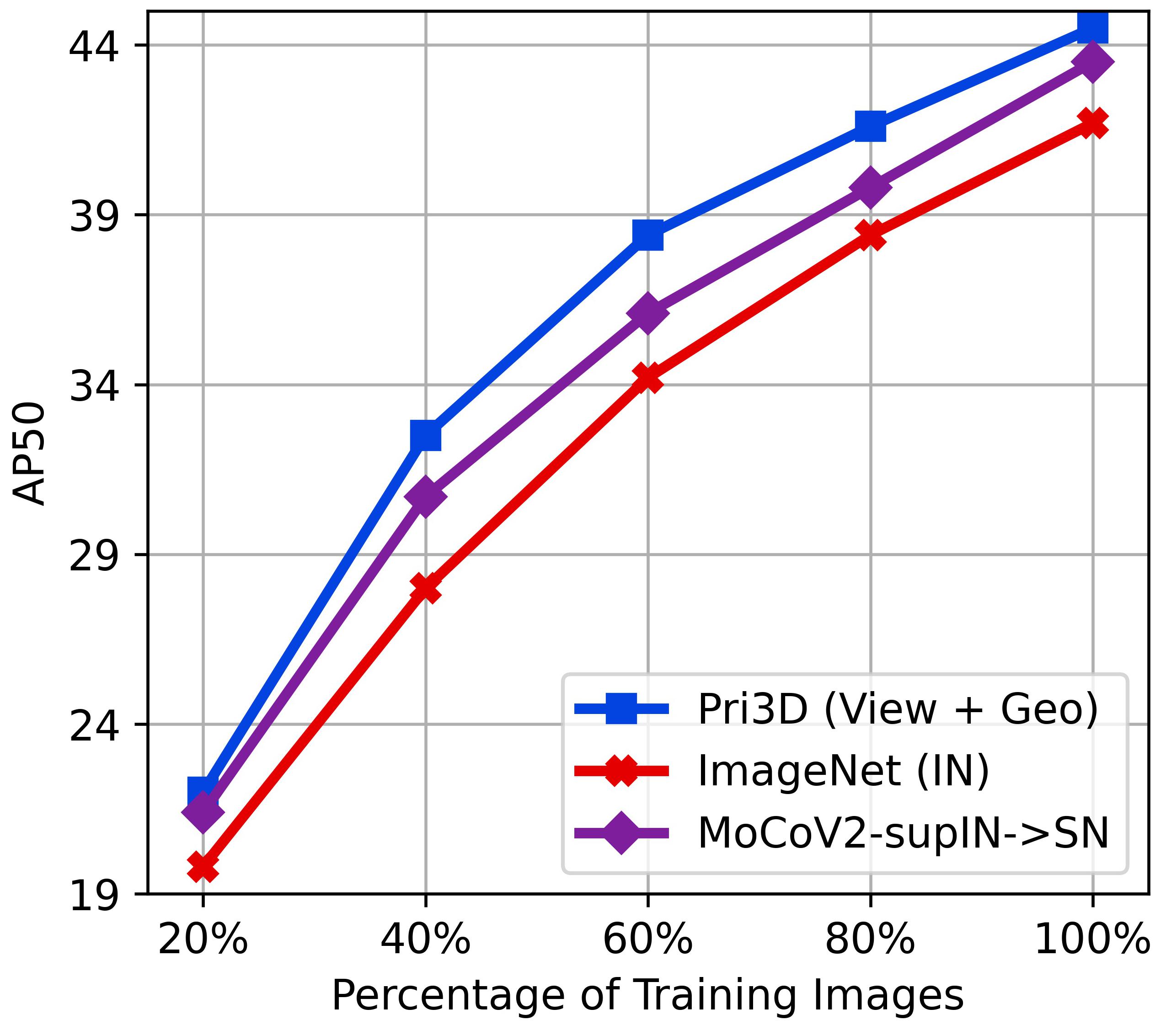}
\caption{\textbf{Data Efficient Learning on ScanNet 2D Detection Task.} Similar to other tasks, our pre-training algorithm achieves consistently better results across various amounts of training data available. The backbone used is ResNet50. }
\label{fig:det_efficient}
\end{figure}

\begin{figure}[h]
\small
\centering
\includegraphics[width=0.9\linewidth]{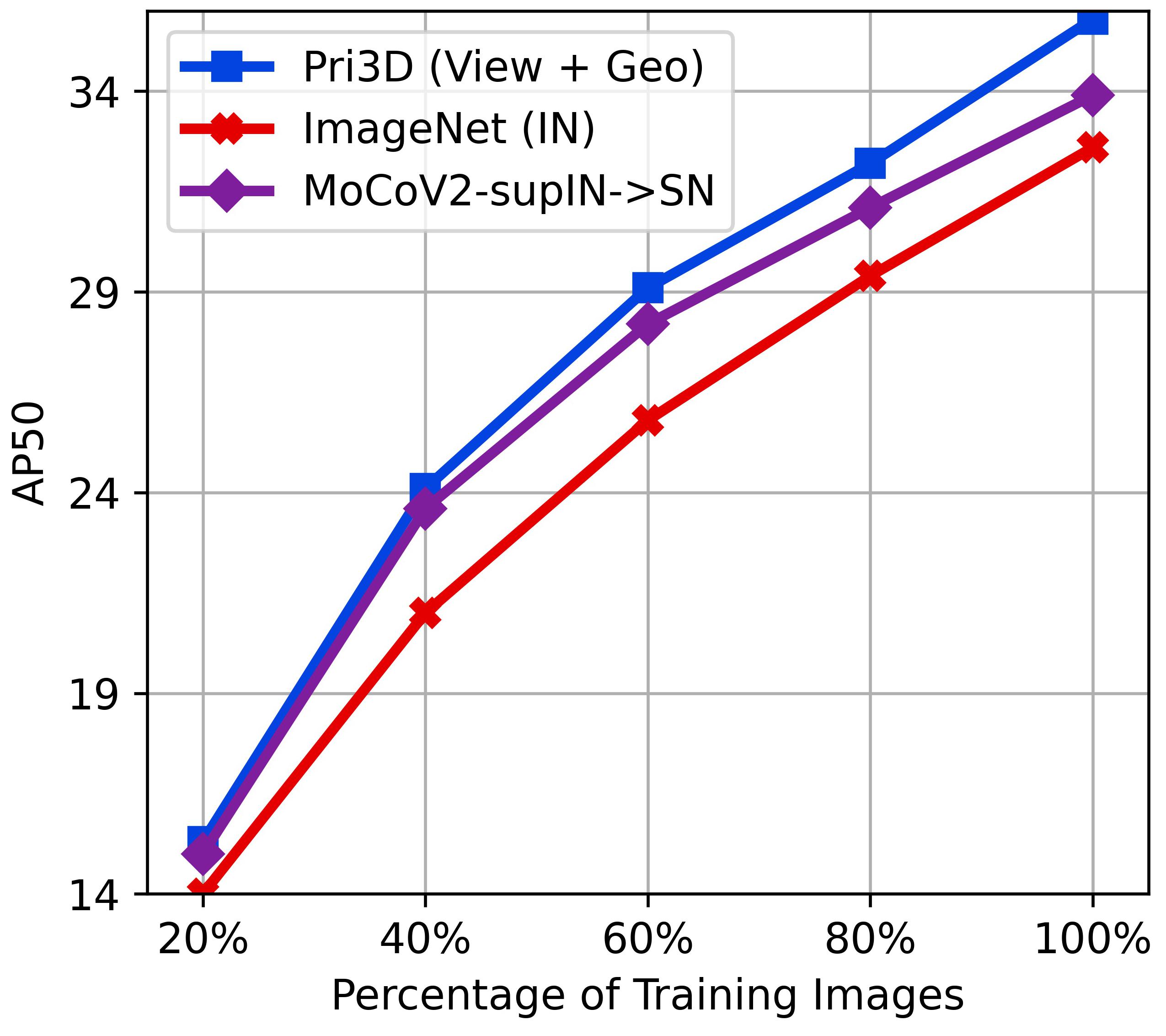}
\caption{\textbf{Data Efficient Learning on ScanNet 2D Instance Segmentation Task.} Our approach provides improved performance across varying amounts of train images available. We use  a ResNet50 backbone. }
\label{fig:insseg_efficient}
\end{figure}

To demonstrate that our pre-training algorithm generalizes well across different downstream tasks in limited data scenarios, we further plot data-efficient learning curves on the 2D object detection task on ScanNet in Figure~\ref{fig:det_efficient}, as well as data-efficient learning curves on the 2D instance segmentation task in Figure~\ref{fig:insseg_efficient}. We use Mask-RCNN~\cite{he2017mask} with a ResNet50 backbone for both tasks.

\section{Additional Qualitative Visualizations}\label{sec:vis}

\begin{figure*}[h!]
	\centering
	\includegraphics[width=1.0\linewidth]{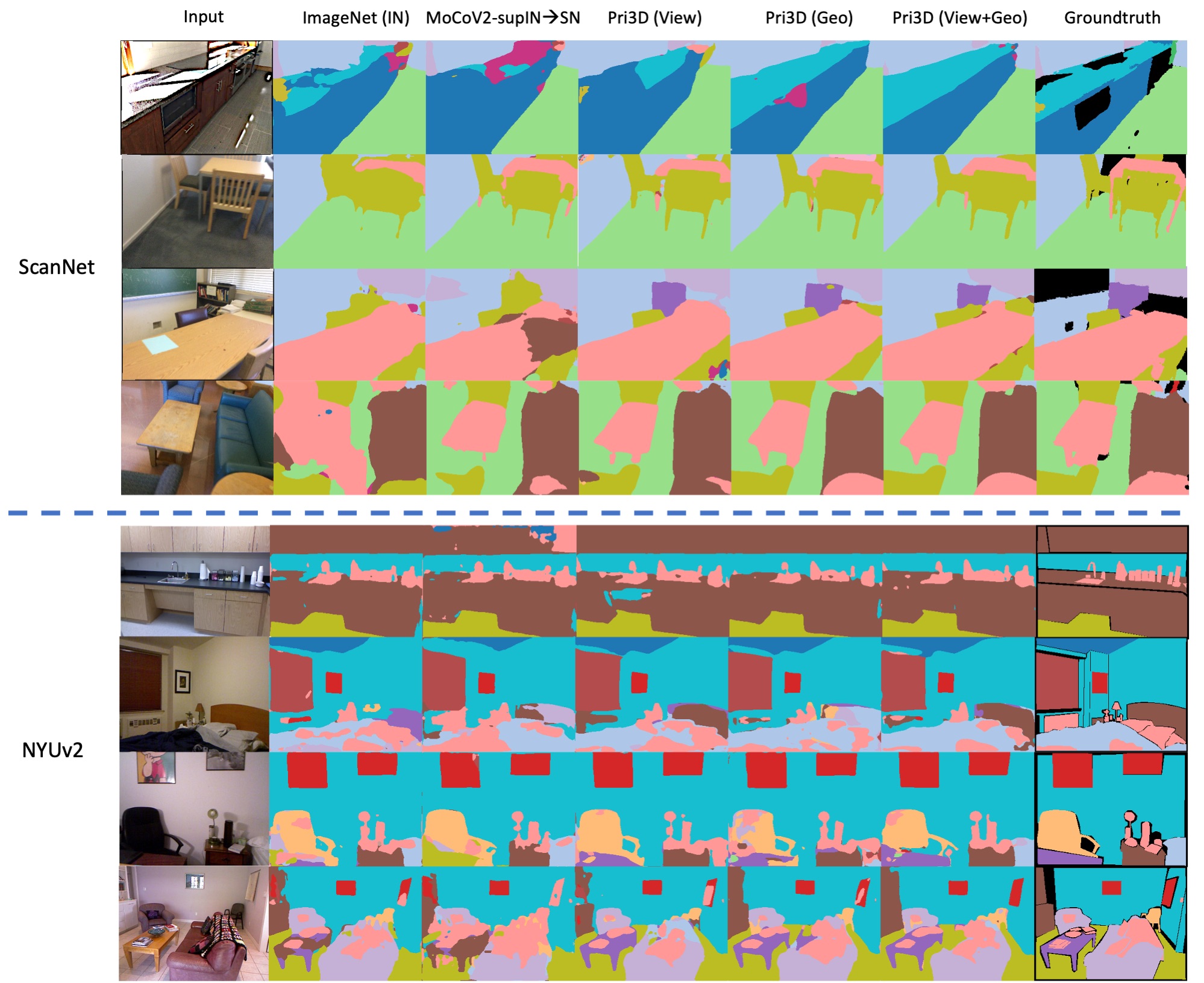}
	\caption{We show qualitative results on 2D semantic segmentation of ScanNet~\cite{dai2017scannet} and NYUv2~\cite{silberman2012indoor}. By encoding 3D priors, we obtain improved segmentation results, particularly for objects that are occluded or have more specular material properties. } 
	\label{fig:quality_result}
\end{figure*}

We additionally show more visualizations of 2D semantic segmentation  on the ScanNet~\cite{dai2017scannet} and NYUv2~\cite{silberman2012indoor} datasets (see Figure~\ref{fig:quality_result}). We can achieve notably improved segmentation results by using our pre-trained weights than ImageNet pre-trained weights.

\section{Generalization Across Datasets.}\label{sec:generalization}
We conduct additional experiments on more diverse datasets for pre-training (ScanNet, MegaDepth~\cite{MegaDepthLi18}, KITTI~\cite{Geiger2012CVPR}) and downstream tasks including COCO~\cite{lin2014microsoft} and outdoor data (Cityscapes~\cite{cordts2016cityscapes} and KITTI) (see Table~\ref{tab:coco}). We still observe a consistent improvement with our pre-training methods across different datasets and tasks. On COCO, the gap is not as significant as it in other scenarios, due to the drastic domain gap between COCO and our 3D dataset for pre-training.

\begin{table}[h]
  \vspace{-0.2cm}
  \centering
   \resizebox{0.45\textwidth}{!}{
  \begin{tabular}{l|c|c|c}
  \specialrule{1.1pt}{0.1pt}{0pt}
Pretraining &  Finetuning & Task & Metric \\
  \hline
   ImageNet-Pretrain        & COCO       & det (mAP)         & 59.5 \\
   Pri3D (ScanNet)      & COCO       & det (mAP)         & \textbf{60.6} \\ \hline
   ImageNet-Pretrain        & COCO       & ins (mAP)      & 56.6 \\
   Pri3D (ScanNet)      & COCO       & ins (mAP)  & \textbf{57.5} \\ \hline
   ImageNet-Pretrain        & Cityscapes & sem (mIoU)  & 54.1 \\ 
   Pri3D (MegaDepth)    & Cityscapes & sem (mIoU)  & 55.2\\ 
   Pri3D (KITTI)        & Cityscapes & sem (mIoU)  & \textbf{56.3} \\ \hline
   ImageNet-Pretrain        & KITTI      & sem (mIoU)  & 28.5 \\ 
   Pri3D (MegaDepth)    & KITTI      & sem (mIoU)  & 30.8 \\ 
   Pri3D (KITTI)        & KITTI      & sem (mIoU)  & \textbf{33.2} \\
  \specialrule{1.1pt}{0.1pt}{0pt}
  \end{tabular}
  }
  \caption{Mask R-CNN is used for detection (det) and instance segmentation (ins); ResUNet50 for semantic segmentation (sem).}
  \vspace{-0.3cm}
\label{tab:coco}
\end{table}

\section{What The Contrastive Loss Learns}\label{sec:correspondence}

\begin{figure*}
	\centering
	\includegraphics[width=0.95\linewidth]{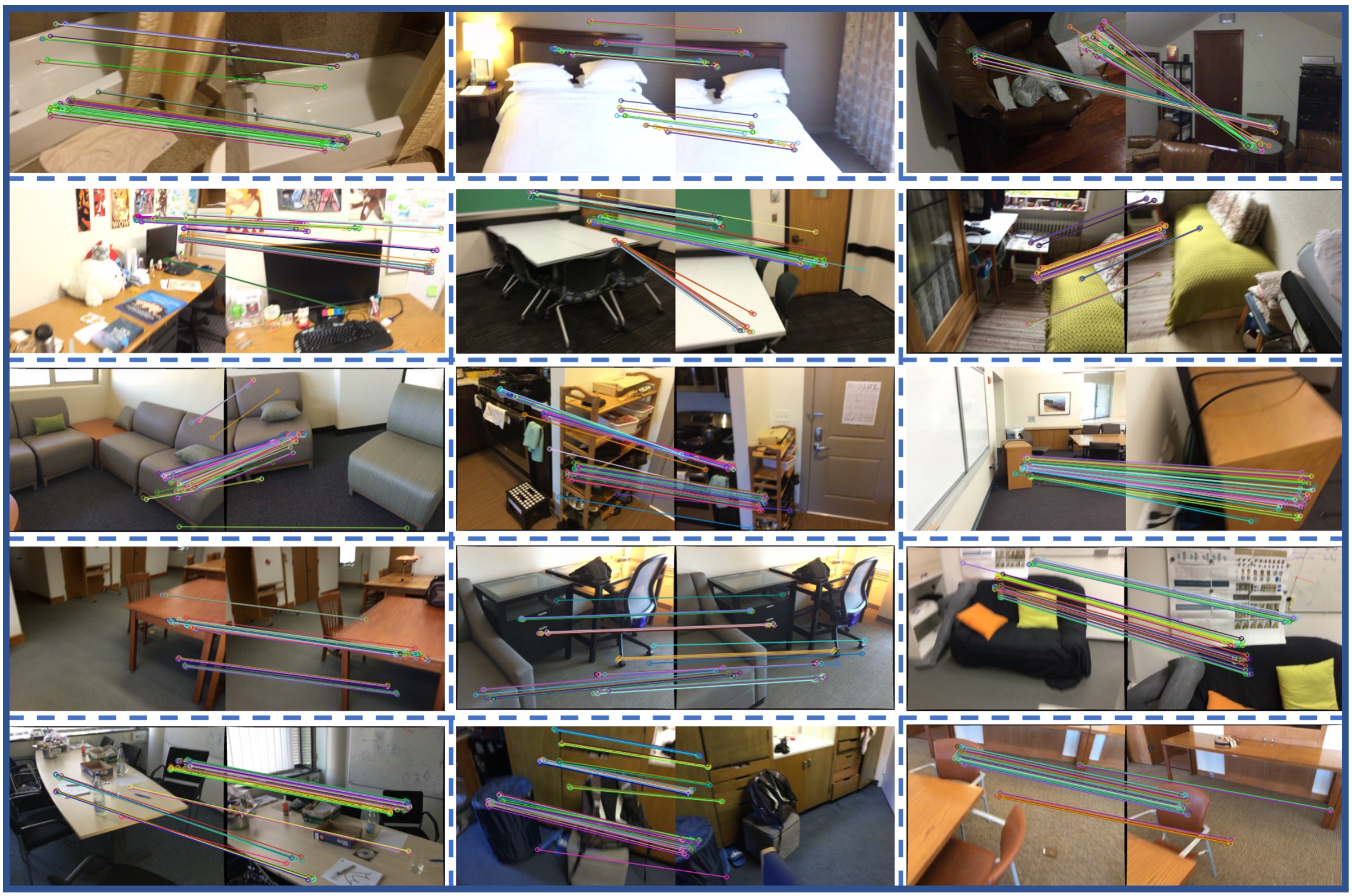}
	\caption{Samples of pixel-level correspondences by matching the features inferred from pre-trained models (View+Geo) with a nearest neighbour search in feature space. Combining 3D priors enables to find correspondences in some extremely difficult cases, such as with large view angle changes and on the surfaces with the same color. }
	\label{fig:matching}
\end{figure*}

To demonstrate our pre-training algorithm has learned meaningful features, we show some visualizations of correspondences matching. We first use pre-trained weights to make a forward pass inference on a pair of images to get features for each pixel. For each pixel in one image, we do nearest neighbour search to find its closed corresponding pixel in anther image in feature space. We draw the matching in this way in Figure~\ref{fig:matching}.

\section{Unsupervised Pre-training Pipeline}
\label{sec:unpri3d}
We experiment with network weights trained with self-supervision on ImageNet for encoder initialization. This is also a two-stage pipeline but without using any semantic labels in either stage. Even though the use of a supervised ImageNet pre-trained initialization is a common practice, for completeness we also evaluate Pri3D in an unsupervised pipeline without using ImageNet labels; we demonstrate the experimental results of \textbf{Unsupervised Pri3D} in Table~\ref{tab:moco}. Results suggest that Pri3D does not rely on any semantic supervision (\eg ImageNet labels) to succeed, and still is able to achieve a substantial gain in this setup. To clarify the baselines:
\begin{itemize}
\item \textbf{Unsupervised ImageNet Pre-training (MoCoV2-IN)}. We use MoCoV2~\cite{chen2020improved} ImageNet pre-trained weights. \emph{No ScanNet data is involved.}
\item \textbf{2-Stage MoCoV2 (MoCoV2-unsupIN$\rightarrow$SN)}. We start with MoCoV2-IN as the encoder initialization, but add another stage to finetune MoCoV2 with randomly shuffled ScanNet images. In this case, \emph{we use ScanNet data but no 3D priors are used.}
\end{itemize}

\begin{table}[h]
  \centering
  \small
  \begin{tabular}{l|c}
  \specialrule{1.1pt}{0.1pt}{0pt}
Method &  mIoU  \\
  \hline
   \textcolor{gray}{Scratch} & \textcolor{gray}{39.1} ~~ \\
   ImageNet Pre-training (IN) & 55.7 ~~ \\
   \hline
   MoCoV2-IN   &\quad 54.6~\tiny{\textcolor{gray}{(-1.1)}} \\
   MoCoV2-unsupIN$\rightarrow$SN &\quad 55.4~\tiny{\textcolor{gray}{(-0.3)}} \\
   \hline
   Unsupervised Pri3D (View) &\quad 60.5~\tiny{\textcolor{gray}{(+4.8)}} \\
   Unsupervised Pri3D (Geo)  &\quad 60.8~\tiny{\textcolor{gray}{(+5.1)}} \\
   Unsupervised Pri3D (View + Geo) &\quad \textbf{60.8~\tiny{\textcolor{darkgreen}{(+5.1)}}} \\
  \specialrule{1.1pt}{0.1pt}{0pt}
  \end{tabular}
  \caption{\textbf{2D Semantic Segmentation on ScanNet (Unsupervised Pre-training Pipeline).}  We additionally show that for Pri3D encoder initialization (stage I), we can replace the ImageNet pre-trained weights with (self-supervised) MoCoV2 weights; the whole pipeline does not require semantic labels. Pri3D still shows a large improvement over supervised ImageNet pre-training and compare favorably with strong MoCo-style baselines. All experiments are with a ResNet50 backbone.}
\label{tab:moco}
\end{table}

\section{Depth Prediction As A Proxy Loss}
\label{sec:depth_proxy}
We explore the influence of 3D priors on 2D tasks starting with the simplest 3D prior, i.e., depth prediction~\cite{Hu2019RevisitingSI}. Depth prediction indicates a positive signal but not strong enough. We then try different other stronger 3D priors, and further use depth prediction as a proxy loss by default. Moreover, we show an ablation study on the depth proxy loss in Table~\ref{tab:no_depth}.

\begin{table}[h]
  \centering
  \small
  \begin{tabular}{l|c}
  \specialrule{1.1pt}{0.1pt}{0pt}
Method &  mIoU  \\
  \hline
   \textcolor{gray}{Scratch} & \textcolor{gray}{39.1}  \\
   ImageNet Pre-training (IN) & 55.7  \\
   Pri3D (View) w/o depth proxy  & 60.2 \\
   Pri3D (View) & 61.3 \\
  \specialrule{1.1pt}{0.1pt}{0pt}
  \end{tabular}
  \caption{\textbf{2D Semantic Segmentation on ScanNet.} All experiments are with a ResNet50 backbone.}
\label{tab:no_depth}
\end{table}

\section{Limitations}
\label{sec:improvements}
While our approach demonstrates the promising effect of learning 3D priors for 2D representation learning, there are various limitations.
For instance, joint 2D and 3D pre-training, in contrast to our current 3D-based constraints only, would likely provide the most informative signal for representation learning for downstream tasks.
Additionally, our current 3D-based pre-training leverages indoor scene data from ScanNet, and we would expect further generalizability by augmentation with data from other environments, such as outdoor scene data (e.g., \cite{geiger2013vision,cordts2016cityscapes}).

\end{appendix}

\end{document}